\documentclass[reprint,notitlepage,longbibliography,amsmath,amssymb,floats,superscriptaddress,nofootinbib,10pt]{revtex4-1}
\usepackage{mymacros}

\begin{document}

\title{Criticality versus uniformity in deep neural networks}

\author{Aleksandar Bukva}
\email{bukva@lorentz.leidenuniv.nl }
\affiliation{Instituut-Lorentz, $\Delta$-ITP, Universiteit Leiden, P.O. Box 9506, 2300 RA Leiden, The Netherlands.}

\author{Jurriaan de Gier}
\affiliation{Instituut-Lorentz, $\Delta$-ITP, Universiteit Leiden, P.O. Box 9506, 2300 RA Leiden, The Netherlands.}

\author{Kevin T. Grosvenor}
\email{grosvenor@lorentz.leidenuniv.nl}
\affiliation{Instituut-Lorentz, $\Delta$-ITP, Universiteit Leiden, P.O. Box 9506, 2300 RA Leiden, The Netherlands.}

\author{Ro Jefferson}
\email{r.jefferson@uu.nl}
\affiliation{Institute for Theoretical Physics, and Department of Information and Computing Sciences, Princetonplein 5, 3584 CC Utrecht, The Netherlands}

\author{Koenraad Schalm}
\email{kschalm@lorentz.leidenuniv.nl }
\affiliation{Instituut-Lorentz, $\Delta$-ITP, Universiteit Leiden, P.O. Box 9506, 2300 RA Leiden, The Netherlands.}

\author{Eliot Schwander}
\affiliation{Instituut-Lorentz, $\Delta$-ITP, Universiteit Leiden, P.O. Box 9506, 2300 RA Leiden, The Netherlands.}

\begin{abstract}
Deep feedforward networks initialized along the edge of chaos exhibit exponentially superior training ability as quantified by maximum trainable depth. In this work, we explore the effect of saturation of the tanh activation function along the edge of chaos. In particular, we determine the line of uniformity in phase space along which the post-activation distribution has maximum entropy. This line intersects the edge of chaos, and indicates the regime beyond which saturation of the activation function begins to impede training efficiency. Our results suggest that initialization along the edge of chaos is a necessary but not sufficient condition for optimal trainability.
\end{abstract}

\maketitle


\emph{Introduction.} Over the past decade or so, deep learning has emerged as one of the most powerful tools for processing and analyzing data, and has proven successful on an increasingly wide range of computational challenges. These remarkable feats include highly accurate image classification \cite{NIPS2012_c399862d}, advanced generative modelling of images \cite{2021arXiv210212092R}, natural language processing \cite{2020arXiv200514165B}, accurate protein structure predictions \cite{Jumper:2021wp}, and besting humans in a wide range of games \cite{Schrittwieser:2020ti}. Key to these neural networks' success is the extremely large number of parameters---generally speaking, the \emph{expressivity} of a neural network increases with depth \cite{2016arXiv160605336R}. Expressivity refers to the range of functions that a network can approximate, with the network being understood as simply a function from the space of inputs to the space of outputs. However, the price we must pay for larger and more powerful networks is that they are more difficult to train; for example, the risk of vanishing or exploding gradients is exacerbated with depth \cite{geron2019hands}. Hence, an improved understanding of how the network parameters impact trainability is highly valuable, as even small improvements in the initialization of deep neural networks can make intractable problems tractable.

In this work, we study trainability in deep random feedforward neural networks. Such networks are frequently used in the literature due to their analytical tractability: the phase space is two-dimensional and parameterized by the variances of the initial  weight and bias distributions: $\sigma_w^2$ and $\sigma_b^2$.\footnote{As is standard in the literature, we restrict to zero-mean networks, as initializing with a small non-zero mean does not qualitatively change our results.} This makes them useful models for investigating general features of deep networks. In particular, we will be concerned with the behavior of the pre- and post-activations, in terms of both their distributions as well as the accuracy of the network on 
a classic image classification task, namely MNIST (numerical digit recognition) and CIFAR-10 (colored images, which we convert to grayscale).

More specifically, we build on previous work \cite{arxiv.1606.05340,2016arXiv161101232S} which demonstrated the presence of an order-to-chaos phase transition in this class of deep networks. Intui\-tively, correlations in the input that we wish to learn are exponentially suppressed with depth in the ordered (analogously, low-temperature) phase, and washed-out by noise in the chaotic (high-temperature) phase; these two phases are characterized by vanishing or exploding gradients, respectively. The boundary between these two phases is a critical line called the \emph{edge of chaos},\footnote{Technically, this should be called the edge of stability, but we will use edge of chaos synonymously with criticality for consistency with the literature.} which is a continuous phase transition characterized by a diverging correlation length $\xi$ for the layer-to-layer two-point function of the neurons. Since the correlation length sets the depth scale at which information can propagate, this theoretically enables networks of arbitrary depth to be trained at criticality (more generally, networks are trainable provided their depth does not exceed the scale set by $\xi$). In other words, the deeper the network, the closer one must lie to the edge of chaos; this was demonstrated in \cite{2016arXiv161101232S} along a slice of parameter space at bias variance $0.05$ and weight variance ranging from 1 to 4, and subsequently generalized/corroborated in, e.g., \cite{arxiv.1806.05393,arxiv.1806.05394,Erdmenger:2021sot}

Several questions naturally arise from the above work. First, given that the network parameters will evolve under training in order to minimize the specified cost function and, in particular, develop interdependencies, why does the choice of initialization have such a decisive effect on network performance?\footnote{In other words, why does the network remain near the initialization regime (e.g., the edge of chaos) as it evolves?} Indeed, it was observed in \cite{Erdmenger:2021sot} that the hidden-layer pre-activation distributions (as quantified by their variance) rapidly approach some asymptotic value within 10 or fewer layers, and then remain relatively unchanged for arbitrarily many additional layers. We corroborate this fact at the level of the post-activation in fig. \ref{fig:postevol} of appendix \ref{app:postevol}.
 
Second, what role does the particular distribution of post-activations in a given layer play in determining network performance? For example, the activation function considered in \cite{2016arXiv161101232S} is hyperbolic tangent, which we adopt henceforth. When $\sigma_{b}^{2} \ll 1$ and $\sigma_{w}^{2} \lesssim 1$, the pre-activations $z$ of the hidden layers are approximately Gaussian-distributed with small variance (cf. \eqref{eq:varrecursion}). In this case, $\tanh(z)\approx z$, so the network behaves like a linear network. These are quite restrictive, being incapable of representing functions whose output data are non-linearly separable and cannot be generated by a combination of linearly separable data. In the opposite extreme, for large values of $\sigma_{w}^{2}$ and $\sigma_{b}^{2}$, the pre-activation variance becomes so large that the post-activation distribution becomes peaked at $\pm1$. In other words, large pre-activation variance saturates the tanh, causing it to behave like a discrete step-function. One expects this regime to also impair trainability, since the gradients on which the backpropagation algorithm depends become vanishingly small everywhere except near the origin.\footnote{Recall that the updates to the weights and biases under gradient descent contain products of the derivatives of the activation functions in all higher layers.} Thus, it seems that one should seek to remain somewhere between these two extremes. Quantifying this is one of the main motivations for the present work.

\begin{figure*}[t!]
\centering
\begin{subfigure}{8.9cm}
\centering
\includegraphics[width=8cm]{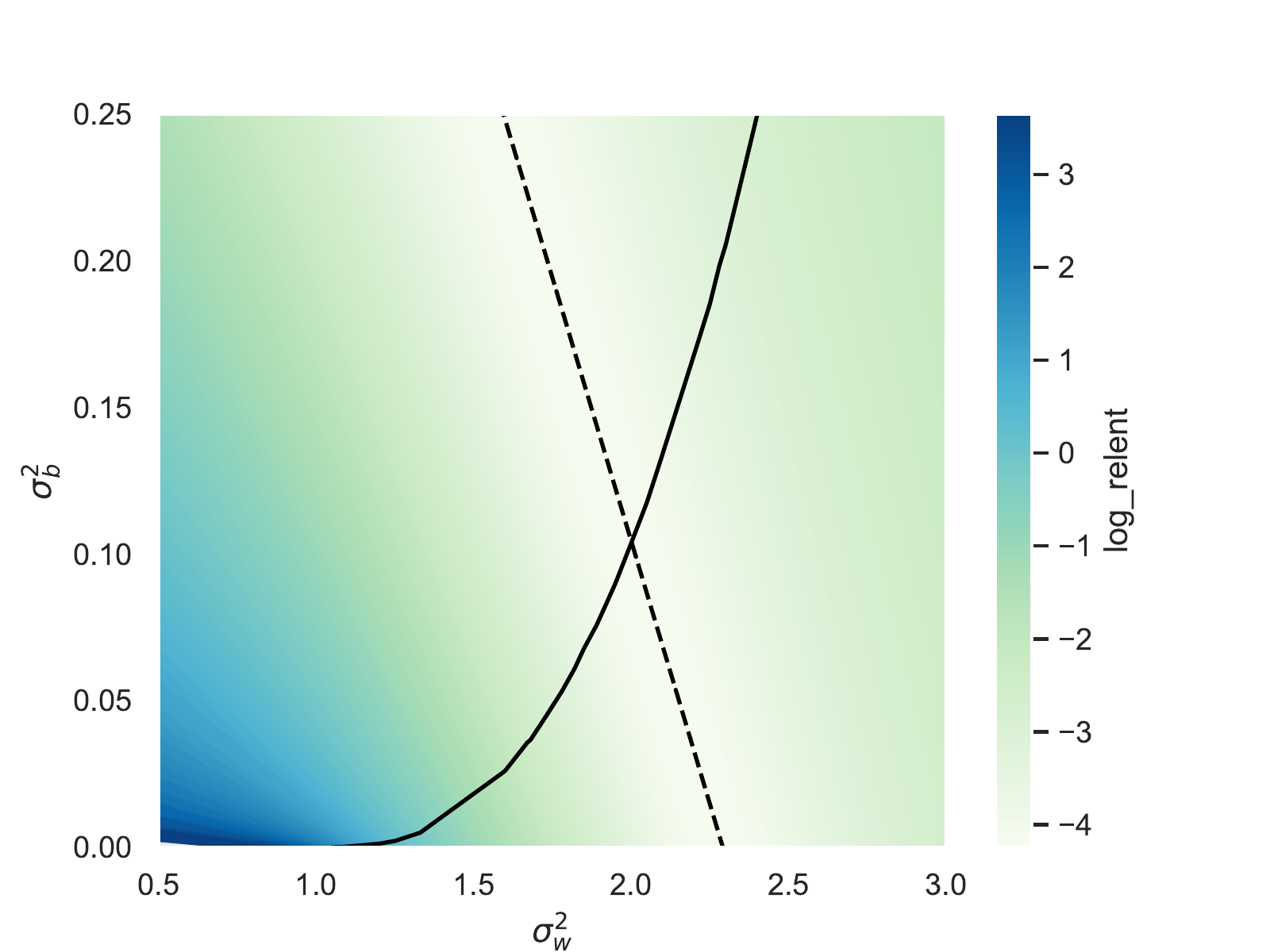}
\end{subfigure}
\begin{subfigure}{8.9cm}
\centering
\includegraphics[width=8cm]{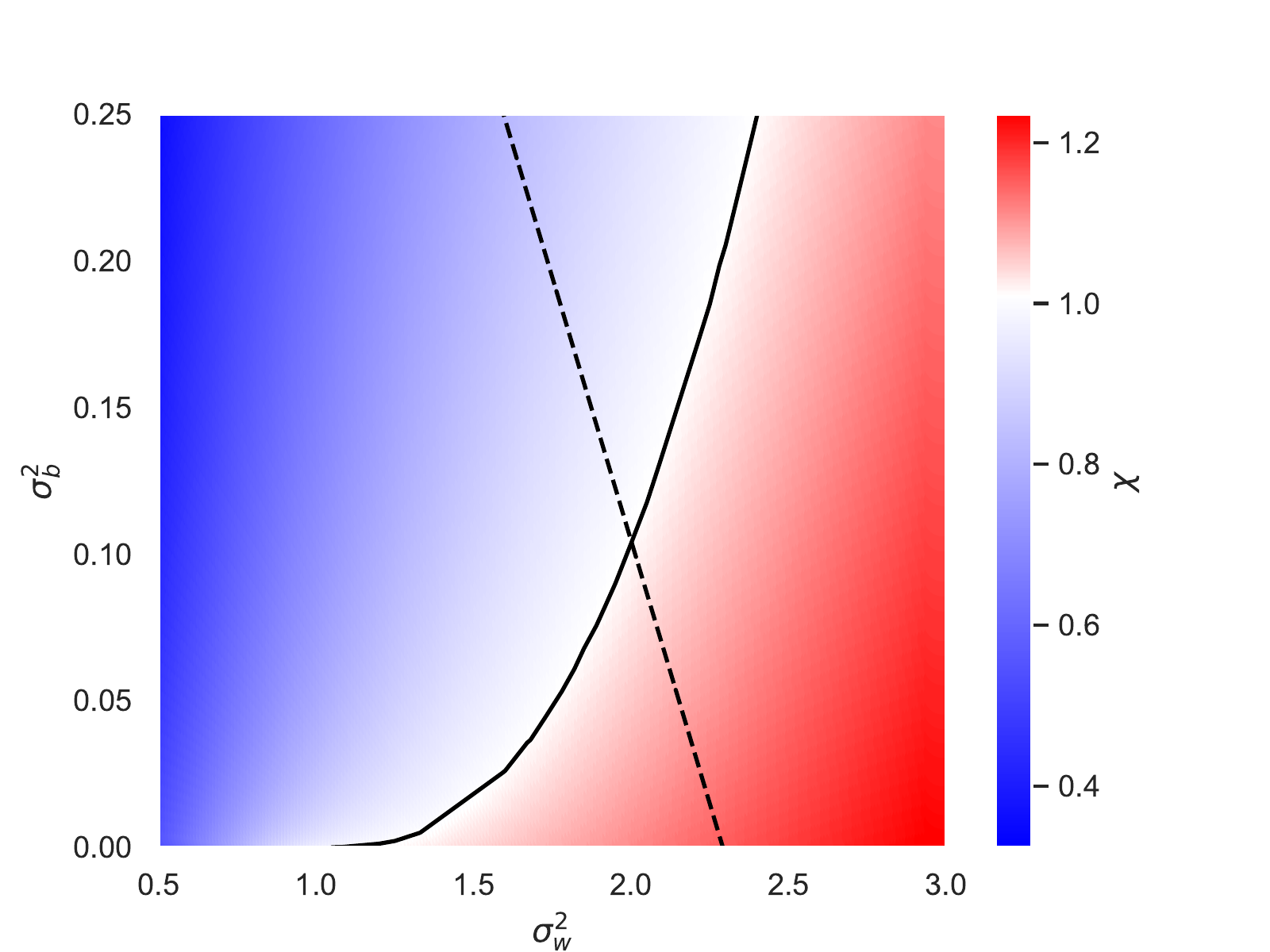}
\end{subfigure}
\caption{(Left) Contour plot of the logarithm of the relative entropy in the $(\sigma_{w}^{2},\sigma_{b}^{2})$ plane. The dashed line is the line of uniformity---saturation increases to the right of it and linearity increases to the left of it. (Right) Contour plot of $\chi=e^{-1/\xi}$. The ordered/low-temperature phase is shaded blue, while the chaotic/high-temperature phase is shaded red. In both, the solid black line is the edge of chaos, while the dashed black line is the line of uniformity.}
\label{fig:loueoc}
\end{figure*}

In particular, note that in both the linear and the saturation regimes, one expects the expressibility of the network to be poor. In contrast, between these extremes lies a region in which the post-activation distribution is approximately uniform, and hence we might expect the expressibility of the network to be maximized at this point. To see this, recall that the uniform distribution has maximum entropy, which measures the number of possible states any particular system can have; a step function, in contrast, can only store a single bit of information, and hence has a low entropy of $\ln2$. This leads to the conjecture that networks whose internal distributions are approximately uniform, i.e., maximally entropic, have higher expressibility, and hence might enjoy a performance advantage. Of course, given approximately Gaussian pre-activations, the post-activation distribution of tanh cannot be exactly uniform, but we can quantify the degree of uniformity via the relative entropy (defined below). In fact, we will show that there is a \emph{line of uniformity} on the $(\sigma_w^2, \sigma_b^2)$ phase space along which the post-activation distribution is as uniform as possible. This line intersects the aforementioned edge of chaos (see fig. \ref{fig:loueoc}), and the relative importance of lying near this line is the primary question we shall explore below. 

We shall begin by deriving an expression for the line of uniformity, defined by the condition that the distribution of the final hidden layer minimizes the relative entropy with respect to the uniform distribution. The computation uses many of the same ingredients as \cite{2016arXiv161101232S}, and the interested reader is encouraged to turn there for more background. We then examine proximity to this line in relation to the edge of chaos considered in previous works. 

\begin{figure*}[t!]
\centering
    \begin{subfigure}{9cm}
    \centering
     \includegraphics[width=8cm]{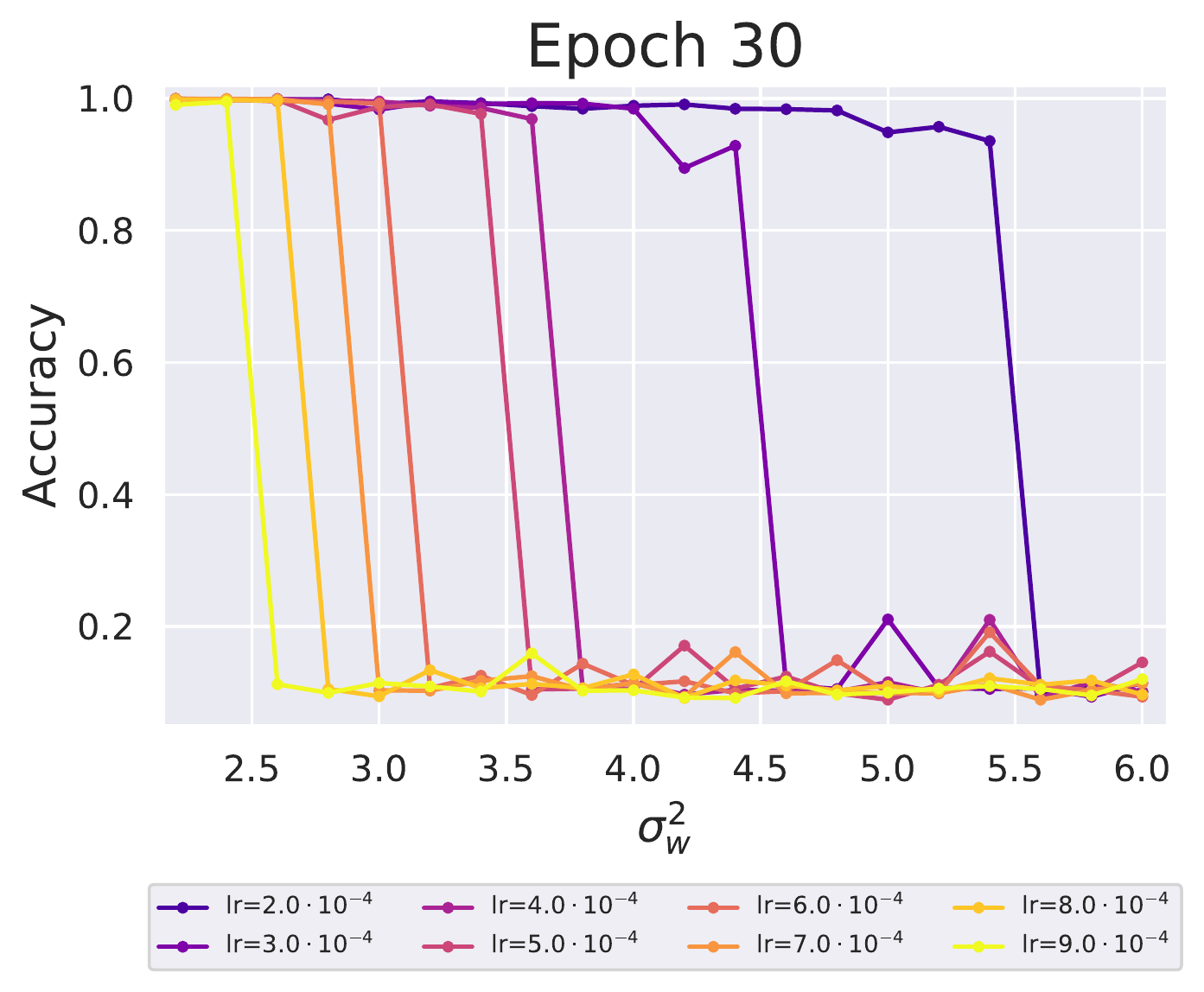}
    \end{subfigure}%
    \begin{subfigure}{9cm}
    \centering
       \includegraphics[width=8cm]{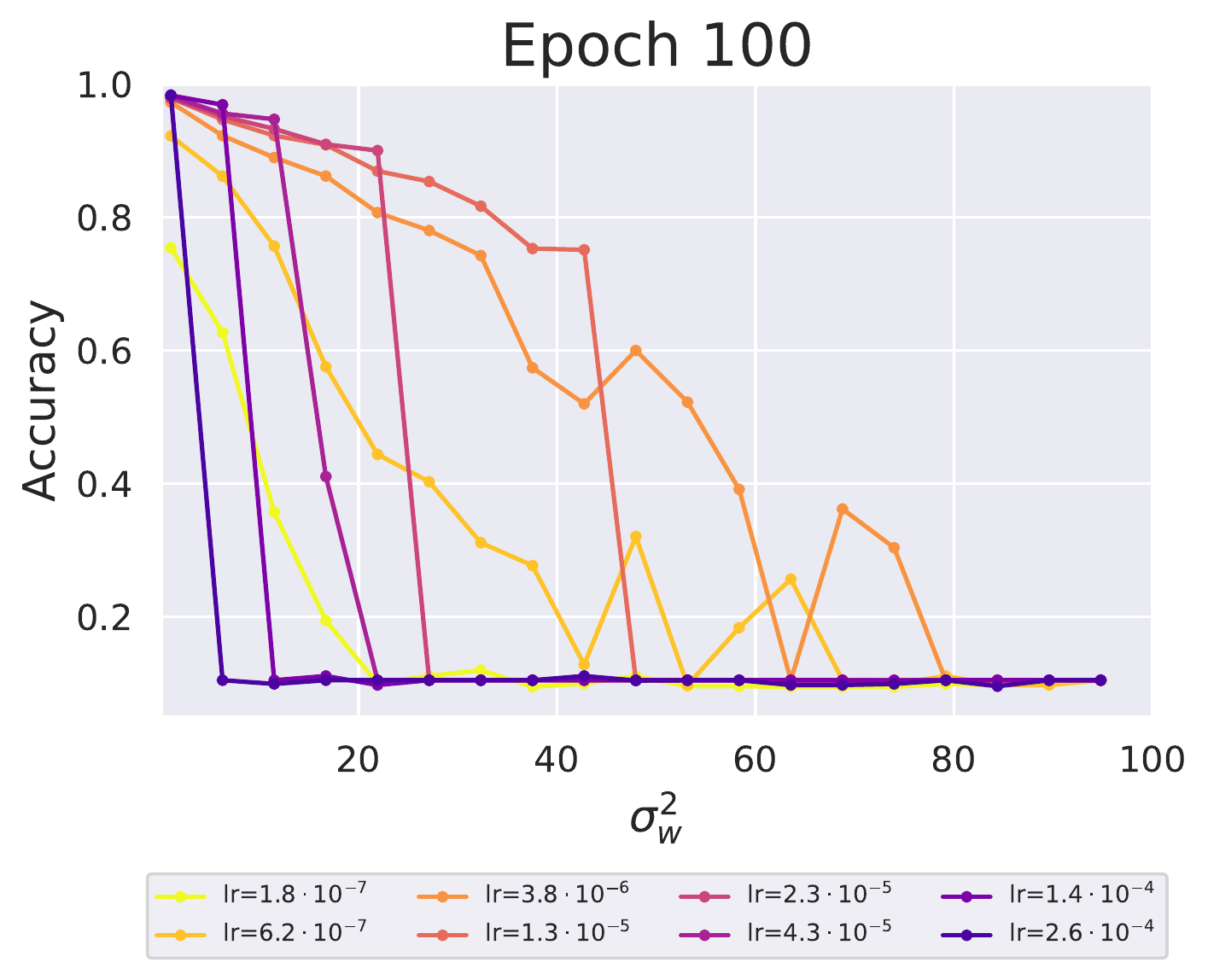}
    \end{subfigure}
    \caption{Accuracies on MNIST for distributions of initial weights along the edge of chaos in a deep ($L=100$), wide ($N=784$) neural network with tanh activation function, for a range of learning rates, after 30 epochs (left) and 100 epochs (right). We observe a drop-off in accuracy beyond a value $\sigma_w^2$ which is up to an order of magnitude larger than the point at which the line of uniformity is crossed. For learning rates of the order typically used in the literature, this point is near the intersection of the LOU and the EOC, but moves to higher values of $\sigma_w^2$ for smaller learning rates. When learning rates become extremely small ($r<10^{-5}$), learning becomes highly inefficient, and the drop-off less sharp for the training duration considered. Networks were trained via stochastic gradient descent with batch size 64 and momentum 0.8.} 
    \label{fig:edge of chaos-drop-off-beyond-lou}
\end{figure*}

We find that for deep networks away from the edge of chaos, the exponential suppression dominates, and no benefit from uniformity is observed. However, along the edge of chaos -- where the suppression is only polynomial -- we find a relatively sharp fall-off in the post-training accuracy to the right of the line of uniformity. The location of this fall-off depends on the learning rate, since decreasing the learning rate can increase the final accuracy, but at the cost of additional computing time (see fig. \ref{fig:edge of chaos-drop-off-beyond-lou}). This suggests that criticality is a necessary but not sufficient condition for optimal trainability. 

This dependence on other hyperparameters illustrates that optimal trainability is not just a matter of final accuracy but also of efficiency, i.e., how quickly the final accuracy is reached. Since computational limits exist, we shall rely on an intuitive notion of efficiency per epochs in addition to accuracy; that is, we consider the accuracy achieved after a fixed number of training epochs. It is conceivable that in the limit of infinite training epochs accuracy differences disappear, so that formally, the configurations are equally good. In a practical sense however, they clearly are not.

Note that there can obviously be very many notions of efficiency depending on which resource(s) one considers most valuable. Here, we are implicitly prioritizing training time, i.e., number of epochs. If one were to put the premium on floating point operations used in training, then one would instead measure efficiency as in \cite{2020arXiv200504305H}. Yet another concept called learning efficiency has to do with how much time it takes to run a learning algorithm and, in particular, how this scales with the size of the input space \cite{2014arXiv1410.1141L}. 

Returning to our main question, to isolate the effects of uniformity \emph{away} from the edge of chaos, we also examine networks which are both shallow (i.e., not yet exponentially suppressed) and narrow (i.e., low expressibility per layer), and confirm that training efficiency, in the sense described above, degrades to the right of the line of uniformity (i.e., away from the origin), though final accuracy need not. In contrast to the edge of chaos, the line of uniformity is not a sharp phase boundary, but it does indicate coarsely the parameter boundary where activation saturation starts to affect training efficiency. This not only establishes the more obvious point that, even in deep random feedforward toy models on the edge of chaos, backpropagation training depends sensitively on activation function choice, as earlier emphasized in \cite{2018arXiv180508266H,pmlr-v97-hayou19a}, but also that for a given activation function choice there are optimal points or regions on the edge of chaos itself. 

\bigskip

\emph{The line of uniformity.} We can estimate the location of the line of uniformity by capitalizing on the fact that wide networks, with a large number $N$ of neurons in each hidden layer, are approximate Gaussian processes. At finite $N$, the neurons in a given layer are not independent due to their shared dependence on the neurons in the previous layer. Physically however, the non-Gaussianities that can be seen by marginalizing over the previous layer(s) can be thought of as interactions that are $1/N$ suppressed \cite{Roberts:2021fes,Grosvenor:2021eol}. Hence, in the limit $N\rightarrow\infty$, the distribution of pre-activations becomes Gaussian, essentially by the central limit theorem. This greatly simplifies the analysis, and is the reason for the widespread use of such models in previous studies, including \cite{2016arXiv161101232S}.\footnote{One will often see the phrase ``mean-field theory'' used in place of the central limit theorem in this context; however, as pointed out in \cite{Grosvenor:2021eol}, this is not technically correct, and mean-field theory does not necessarily correspond to the $N\rightarrow\infty$ limit.}

Thus, at large-$N$, the distribution of pre-activations $z$ for any hidden layer takes the form
\begin{equation} \label{eq:pre}
    p (z ; \sigma^2 ) = \frac{1}{\sqrt{2 \pi} \, \sigma} \, e^{- \frac{z^2}{2 \sigma^2}}~,
\end{equation}
where $\sigma^2$ is the variance, and we assume the mean $\mu=0$ since adding a small finite mean does not qualitatively change our results. If the activation function $\phi (z)$ is one-to-one and once-differentiable, then the distribution of post-activations $x$ will be given by
\begin{equation} \label{eq:postgeneral}
    p_{\phi} (x ; \sigma^2 ) = \frac{1}{\sqrt{2 \pi} \, \sigma\,\phi' \bigl( \phi^{-1} (x) \bigr)} \, e^{- \frac{\phi^{-1} (x)^2}{2 \sigma^2}}~.
\end{equation}
Concretely, for $\phi (z) = \tanh (z)$, this yields
\begin{equation} \label{eq:ppost}
    p_{\phi} (x; \sigma^2 ) = \frac{1}{\sqrt{2 \pi} \, \sigma (1 - x^2 )} \, e^{- \frac{\rm{arctanh} (x)^2}{2 \sigma^2}}~,
\end{equation}
with $x\in[-1,1]$. The corresponding variance is given by
\begin{equation} \label{eq:varpost}
	\sigma_{\phi}^{2} = \int_{-1}^{1}\mathrm{d} x \, x^2 \, p_{\phi} \bigl( x ; \sigma^2 \bigr)~.
\end{equation}

As mentioned above, we quantify the uniformity of the post-activation distribution $p_\phi$ by the relative entropy or Kullback-Leibler divergence with respect to the uniform distribution $p_\mathrm{uni}$,
\begin{equation}
S(p_\mathrm{uni}||p_\phi)
=\int_{-1}^1\!\mathrm{d} x\;p_\mathrm{uni}(x)\ln\frac{p_\mathrm{uni}(x)}{p_\phi(x)}~.
\end{equation}
Substituting in \eqref{eq:ppost} and $p_\mathrm{uni}=\tfrac{1}{2}$, this yields
\begin{equation} \label{eq:relent}
    S( p_{\rm{uni}} || p_{\phi} ) = \frac{1}{2}\ln(8\pi\sigma^2)+\frac{\pi^2}{24 \sigma^2}-2~.
\end{equation}
This has a minimum at
\begin{equation} \label{eq:sigmamin}
    \sigma_{\rm{min}}^{2} = \frac{\pi^2}{12} \approx 0.822~.
\end{equation}

Therefore, we wish to find the set of points $(\sigma_w^2,\sigma_b^2)$ at which the variance of the final hidden layer is $\sigma_\mathrm{min}^2$; this  will define the line of uniformity. To proceed, we use the recursion relation
\begin{align} \label{eq:varrecursion}
     \sigma_{\ell}^{2} = \sigma_{w}^{2} \, \sigma_{\phi , \ell -1}^{2} + \sigma_{b}^{2}~,
\end{align}
which follows from the large-$N$ condition discussed above (i.e., the neurons on any given layer can be treated as i.i.d. random variables). Note that this is exactly the same as eq. (3) of \cite{2016arXiv161101232S}, where our $\sigma_{\ell}^{2}$ is their $q_{aa}^{\ell}$ and our $\sigma_{\phi, \ell -1}^{2}$ is the corresponding integral expression.\footnote{Explicitly, the variance can be written as $\sigma_{\phi}^{2} = \int \mathcal{D} z \, \bigl[ \phi ( \sigma z ) \bigr]^2$, where $\mathcal{D}z=\tfrac{\mathrm{d} z}{\sqrt{2\pi}}e^{-\frac{z^2}{2}}$ is the standard Gaussian measure.\label{ft:gauss}} This recursion relation ostensibly requires the variance of the first hidden layer, $\sigma_1^2$, as an input. However, it turns out that \eqref{eq:varrecursion} quickly converges to a fixed value $\sigma_{*}^{2}$, which (by definition) is a function of $\sigma_{w}^{2}$ and $\sigma_{b}^{2}$, but not of $\sigma_{1}^{2}$:
\begin{equation} \label{eq:sigmastar}
    \sigma_{*}^{2} = \sigma_{w}^{2} \, \sigma_{\phi , *}^{2} + \sigma_{b}^{2},
\end{equation}
where $\sigma_{\phi, *}^{2}$ is $\sigma_{\phi}^{2}$ evaluated at $\sigma_{*}^{2}$; see \cite{arxiv.1606.05340} for further discussion of this convergence. In appendix \ref{app:postevol}, we have demonstrated numerically that the corresponding post-activation distribution indeed converges rapidly to one which depends only on the initialization point $(\sigma_w^2, \sigma_b^2)$.

Now, consider a fixed value of $\sigma_*^2$ (and hence also of $\sigma_{\phi , *}^{2}$). Then we can consider \eqref{eq:sigmastar} as an expression for $\sigma_b^2$ as a function of $\sigma_w^2$, which defines a line in phase space of the form
\begin{equation} \label{eq:fixedsigmastar}
    \sigma_{b}^{2} = \sigma_{*}^{2} - \sigma_{\phi , *}^{2} \, \sigma_{w}^{2}.
\end{equation}
where $\sigma_*^2$ is the $y$-intercept, and $- \sigma_{\phi , *}^{2} $ is the slope. Since the relative entropy \eqref{eq:relent} of the final hidden layer is only a function of its variance, the lines of constant $\sigma_*$ given by \eqref{eq:fixedsigmastar} are also lines of constant relative entropy. In particular, the line of uniformity (minimum relative entropy) is given by \eqref{eq:fixedsigmastar} with $\sigma_{*}^{2} = \sigma_{{\rm min}}^{2} = \frac{\pi^2}{12}$, cf. eq. \eqref{eq:sigmamin}. There is no closed-form expression for $\sigma_{\phi,\mathrm{min}}^{2}$, but we can evaluate \eqref{eq:varpost} numerically to obtain $\sigma_{\phi , {\rm min}}^{2} \approx 0.359$. In summary, the line of uniformity (LOU) is given by
\begin{equation} \label{eq:lou2}
    {\rm LOU}: \quad \sigma_{b}^{2} = \sigma_{{\rm min}}^{2} - \sigma_{\phi, {\rm min}}^{2} \, \sigma_{w}^{2},
\end{equation}
with $\sigma_{{\rm min}}^{2} = \frac{\pi^2}{12} \approx 0.822$ and $\sigma_{\phi, {\rm min}}^{2} \approx 0.359$. In the left panel of fig. \ref{fig:loueoc}, we present a contour plot of the logarithm of the relative entropy. The line of uniformity is the dashed black line---to the left of it, as one approaches the origin, is the linear regime; and to the right, the activation becomes more and more saturated. For comparison, the edge of chaos is the solid black line.

\bigskip


\emph{The edge of chaos.} The method for computing the edge of chaos as a function of $\sigma_{w}^{2}$ and $\sigma_{b}^{2}$ is described in \cite{arxiv.1606.05340,2016arXiv161101232S}. Once we have $\sigma_{*}^{2}$, as described previously, then we can define the quantities
\begin{align} \label{eq:corlength}
    \chi &= \sigma_{w}^{2} \int \mathcal{D} z \, \bigl[ \phi ' ( \sigma_* z ) \bigr]^2, &%
    \xi &= - \frac{1}{\ln \chi},
\end{align}
where $\mathcal{D} z$ is the standard Gaussian measure, cf. footnote \ref{ft:gauss}, and $\xi$ is the correlation length mentioned in the introduction (note that this is denoted $\xi_c$ in \cite{2016arXiv161101232S}). 

The meaning of $\chi$ will be discussed in the next paragraph, while the meaning of $\xi$ is as follows: we consider two identical copies of the network and feed them slightly different inputs. Then, we can study the correlation (i.e., covariance) between a neuron in one copy and the same neuron in the second copy as a function of the layer. This correlation will decay exponentially for deeper layers with a characteristic length scale, $\xi$. (Strictly speaking, this is only true in the ordered phase: in the chaotic phase, the quantity $\xi$ is complex-valued and cannot be interpreted as a correlation length). The edge of chaos is defined as the critical point, where the correlation length $\xi$ diverges.

As discussed in more detail in \cite{arxiv.1606.05340, 2016arXiv161101232S}, $\chi$ is obtained as the derivative of the aforementioned covariance with respect to that in the previous layer, and probes the stability of the fixed point when the covariance is unity: $\chi>1$ implies that we approach this point from below (unstable), while $\chi<1$ implies that we approach this point from above (stable).\footnote{See \cite{roCrit} for a pedagogical explanation.} The edge of chaos corresponds to $\chi=1$, where $\xi$ diverges.  

To find the edge of chaos, we can scan over the space of tuples $(\sigma_w, \sigma_*)$ to find those which satisfy the condition $\chi=1$. We then feed these into \eqref{eq:varrecursion} to find the corresponding value of $\sigma_b$. In this manner, we can find arbitrarily many points on the edge of chaos (EOC). Within some finite range of $\sigma_{w}^{2}$ values, we can find a good fit to the EOC. In the range $1 \leq \sigma_{w}^{2} \leq 10$, a good polynomial fit is
\begin{align} \label{eq:eocfit}
    {\rm EOC}: \quad \sigma_{b}^{2} &= \sum_{n=2}^{9} \frac{c_n}{n!} ( \sigma_{w}^{2} - 1 )^n,
\end{align}
with fit coefficients
\begin{equation}
\setlength{\tabcolsep}{5pt}
\renewcommand{\arraystretch}{1.2}
\begin{tabular}{ll|ll}
\hline
$n$ & $\phantom{-} c_n$ & $n$ & $\phantom{-} c_n$ \\
\hline\hline
2 & $\phantom{-} 0.0190$ & 6 & $-1.15$ \\
3 & $\phantom{-} 0.778$ & 7 & $\phantom{-} 0.769$ \\
4 & $-1.07$ & 8 & $-0.328$ \\
5 & $\phantom{-} 1.25$ & 9 & $\phantom{-} 0.0672$ \\
\hline
\end{tabular}
\end{equation}
Of course, we can reduce the number of fit coefficients needed by reducing the range of $\sigma_{w}^{2}$ values over which we require the fit to be good.

The form of this fit is designed such that it contains the point $( \sigma_{w}^{2} , \sigma_{b}^{2} ) = (1,0)$, and that the edge of chaos has zero slope at this point. We justify these conditions analytically in appendix \ref{app:analytics}. In the right plot in fig. \ref{fig:loueoc}, we present a contour plot of $\chi$. Again, the edge of chaos is drawn as a solid black line and the line of uniformity as a dashed line. The point of intersection of the edge of chaos and line of uniformity is found to be
\begin{equation} \label{eq:intersect}
    ( \sigma_{w}^{2} , \sigma_{b}^{2} )_{{\rm intersect}} = (2.00, 0.104)~.
\end{equation}

\bigskip


\emph{The impact of uniformity along the edge of chaos.}
To the right of the line of uniformity, neurons begin to saturate the tanh activation function, i.e., approach $\pm1$. This implies that backpropagation based on gradient descent should be less efficient, and hence networks should reach a lower accuracy in a fixed amount of training time. The Google Brain collaboration has already established that at the edge of chaos, learning accuracy is enhanced due to polynomial rather than exponential decay of correlations as a function of network depth \cite{2016arXiv161101232S}. Combining the two insights, optimal learning should therefore take place on the edge of chaos near the line of uniformity. 

To test this hypothesis, we have performed the MNIST image classification task in networks ranging up to a depth of $L=100$ hidden layers at various points along the edge of chaos. The resulting learning accuracy is shown in fig.~\ref{fig:edge of chaos-drop-off-beyond-lou}. We see that this expectation is partially validated. On the left side of the line of uniformity -- but to the right of the linear regime -- all points on the edge of chaos are equally good at learning. But beyond a certain point, which lies to the right of the intersection point \eqref{eq:intersect} of the edge of chaos and line of uniformity, the final accuracy decreases. However, this drop-off point is substantially (up to an order of magnitude) displaced to the right of the intersection point, indicating that the line of uniformity is perhaps better thought of as a region rather than a narrow band, and depends on hyperparameters (such as the learning rate) as mentioned above. Nevertheless, for typical learning rates used in the literature of order $10^{-3}$, such as used in \cite{2016arXiv161101232S}, the drop-off point at approximately $\sigma_{w}^{2} \sim 2.5$ is indeed fairly close to the intersection between the line of uniformity and the edge of chaos at $\sigma_{w}^{2} = 2$. 

\begin{figure*}[t!]
\centering
    \begin{subfigure}{9cm}
    \centering
     \includegraphics[width=8cm]{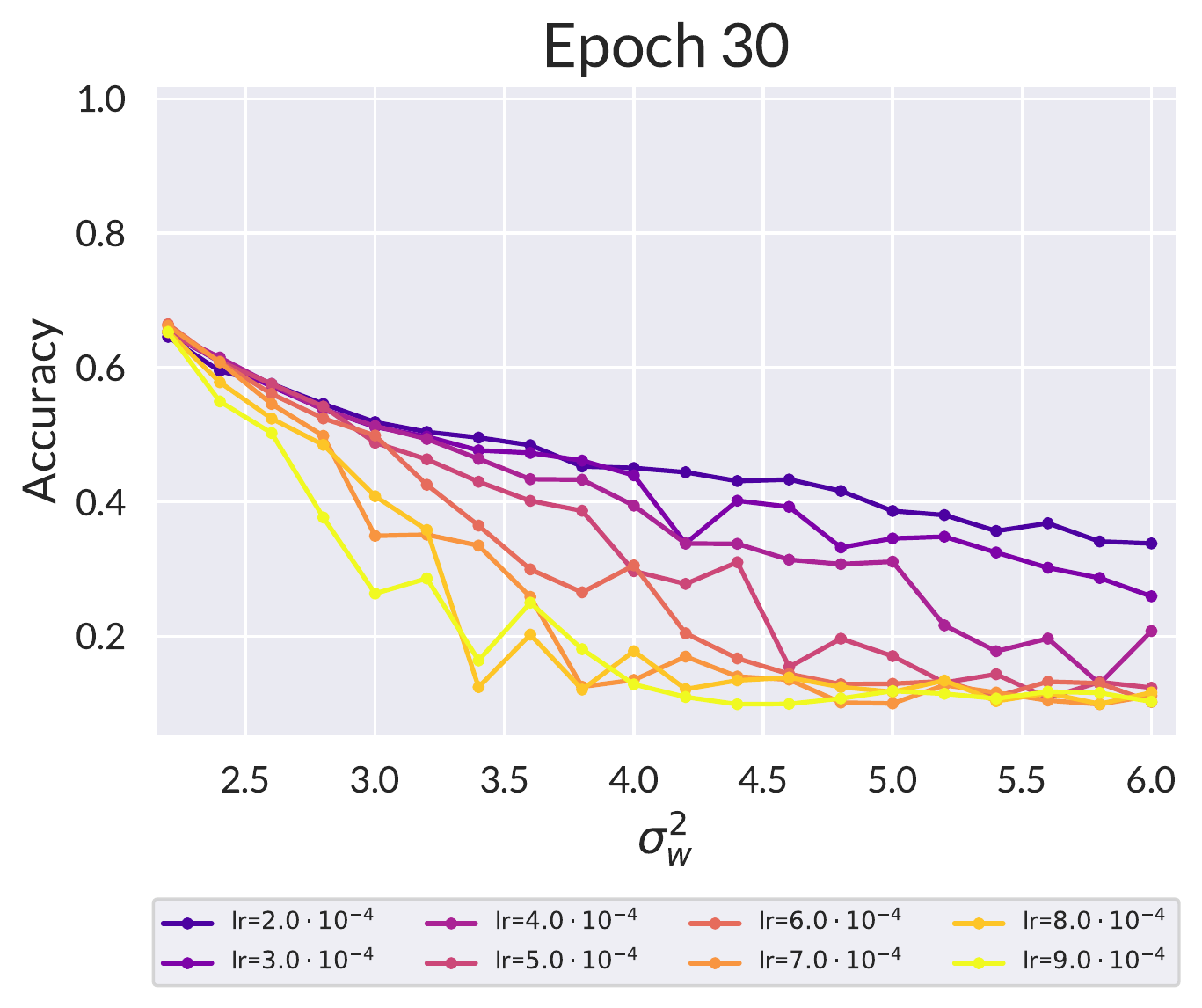}
    \end{subfigure}%
    \begin{subfigure}{9cm}
    \centering
       \includegraphics[width=8cm]{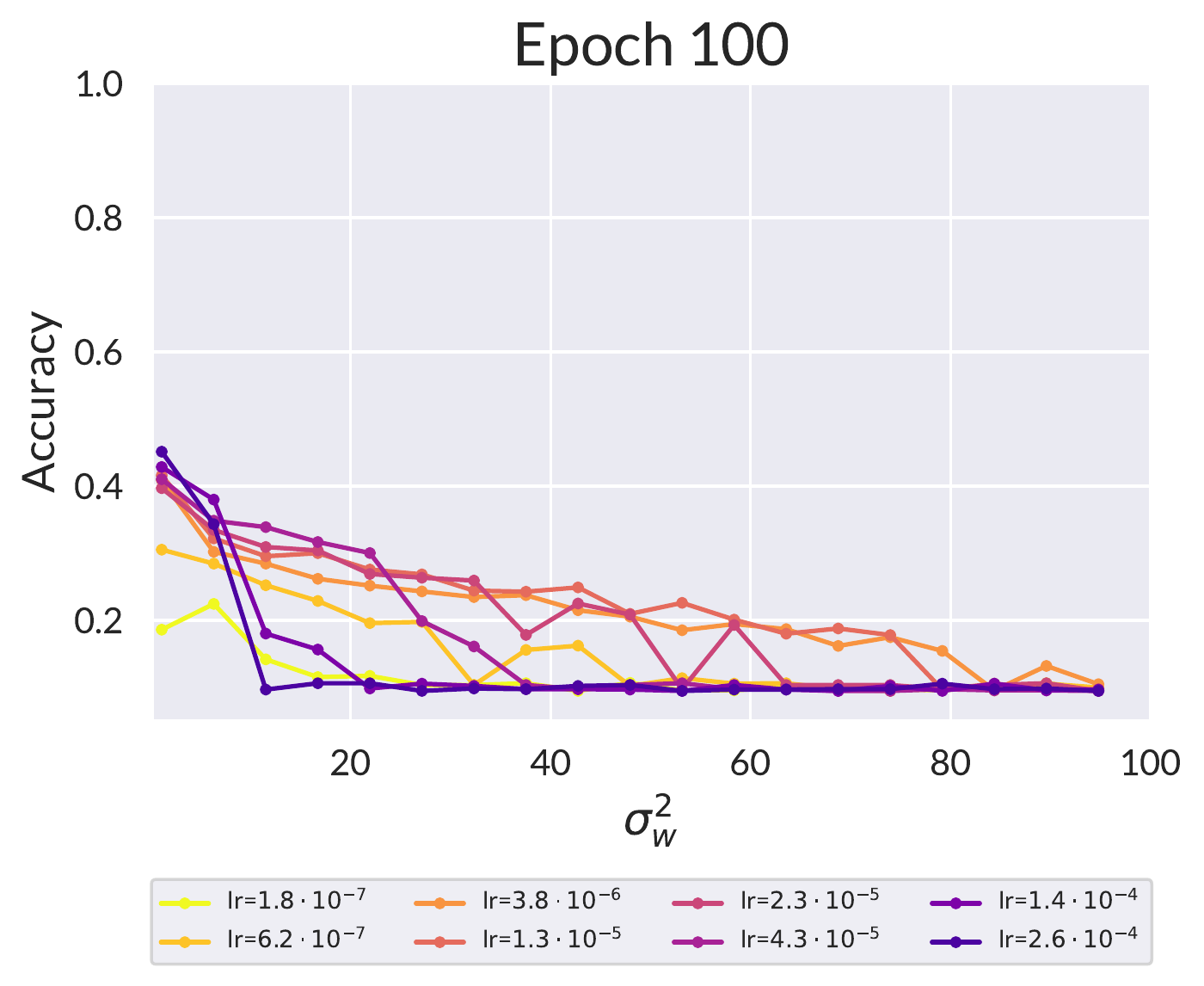}
    \end{subfigure}
    \caption{Accuracies on CIFAR-10 for distributions of initial weights along the edge of chaos in a deep ($L=100$), wide ($N=1024$) neural network with tanh activation function, for a range of learning rates, after 30 epochs (left) and 100 epochs (right). The drop-off in accuracy towards higher values of $\sigma_{w}^{2}$ here is much more gradual than the sharp drop-offs observed for MNIST in fig. \ref{fig:edge of chaos-drop-off-beyond-lou} (at all but the lowest learning rates, $r < 10^{-5}$). Networks were trained via stochastic gradient descent with batch size 64 and momentum 0.8.} 
    \label{fig:EOCdropoffCIFARtanh}
\end{figure*}

We repeated this exercise for the CIFAR-10 image classification task, and present the corresponding results in fig. \ref{fig:EOCdropoffCIFARtanh}. We converted the colored images to grayscale to reduce the input size by a factor of 3. The drop-off in accuracy along the edge of chaos towards larger values of $\sigma_{w}^{2}$ is still present, though the effect is not as dramatic as it is for MNIST. This is not surprising as CIFAR is a much more difficult task than MNIST and so we expect that the saturation of slightly more or fewer neurons will have a much less decisive effect. We note however that in the regime of extremely small learning rates, where training MNIST becomes highly inefficient, the MNIST and CIFAR results appear similar insofar as neither exhibits the obvious sharp drop-off observed for MNIST at the higher learning rates generally used in practice.


Thus, the line of uniformity is not a sharp boundary, unlike the edge of chaos. This is somewhat inherent in its definition, which selects proximity to the uniform distribution of final hidden layer weights as a condition for efficient learning based on the entropic argument given above, but does not specify any particular fall-off behavior. The line of uniformity does, however, give an estimate of where the saturation of the activation function should start to affect learning, and by extension, the point at which saturation of the activation function begins to hinder learning efficiency. To summarize: on the left side of the line of uniformity, the distributions are sufficiently narrow that saturation of the tanh activation function does not occur, and all initial weight distributions along the EOC learn equally well. Conversely, on the right side of uniformity, neurons saturate the activation function and hence hamper learning, even along the EOC. This is our main observation. Importantly, we note that the studies by \cite{2016arXiv161101232S} were performed to the left of the point where the line of uniformity crosses the edge of chaos and hence at optimal efficiency. 

Before moving on to our final set of experiments, we note that the above conclusion is of course specific to saturating activation functions, specifically tanh. This is one motivation for the use of non-saturating activation functions such as ReLU or SWISH, though the unbounded nature of such functions presents its own set of training difficulties. While a similar analysis of uniformity, as quantified by the maximally entropic distribution, for non-saturating activation functions is beyond the scope of this work, a brief inspection of learning efficiency along the EOC for SWISH shows no loss of accuracy in agreement with the absence of saturation effects; see appendix \ref{app:swish}.\footnote{For both SWISH and tanh, the edge of chaos is a line of critical initalizations through phase space, while for ReLU it is only a single point \cite{Roberts:2021fes}.}

\bigskip


\emph{Uniformity away from the EOC.} Thus far, we have examined the impact of uniformity on training efficiency along the edge of chaos. Now, we would like to explore whether the line of uniformity still affords training advantages even for networks initialized far from criticality. In attempting to exhibit this however, one quickly finds that the edge of chaos represents a far more dominant effect than the line of uniformity. A close inspection of the learning accuracy of deep (L=300) and wide (N=784) MNIST learning networks shows that there is no discernible difference in learning accuracy away from the edge of chaos: it is simply poor everywhere (see fig. \ref{fig:googleBrain-repro} in appendix~\ref{app:figs}, also \cite{2016arXiv161101232S}.) This can be understood from the form of the correlation functions: away from the edge of chaos, correlations damp exponentially $\sim e^{-L/\xi}$. For a deep network, this exponential damping will erase any finer difference in accuracy results. Along the edge of chaos, the damping is only polynomial and, therefore, the finer difference remains, as seen in fig.~\ref{fig:edge of chaos-drop-off-beyond-lou}. In shallow networks however, the exponential damping does not have sufficient time to compound, and if the network is also narrow and hence has low expressibility per layer, we can explore the effect of uniformity even away from criticality in such models.

Furthermore, it is common lore that efficient backpropagation needs sufficient gradients, and that such gradients are absent if most of the post-activation functions saturate to a fixed asymptotic value. However, if a sufficient number of weight and bias values are such that there remain trainable paths through a saturated landscape, the model will still learn, even though, distribution-wise, most of the neurons have saturated. Therefore, the inefficiency due to saturation discussed above
can be displayed more clearly  
by choosing narrower networks with smaller $N$, where we might expect that uniformity -- that is, maximally entropic distributions -- may afford the most advantage.

\begin{figure*}[t!]
    \centering
    \begin{subfigure}[t]{9cm}
         \includegraphics[width=8cm]{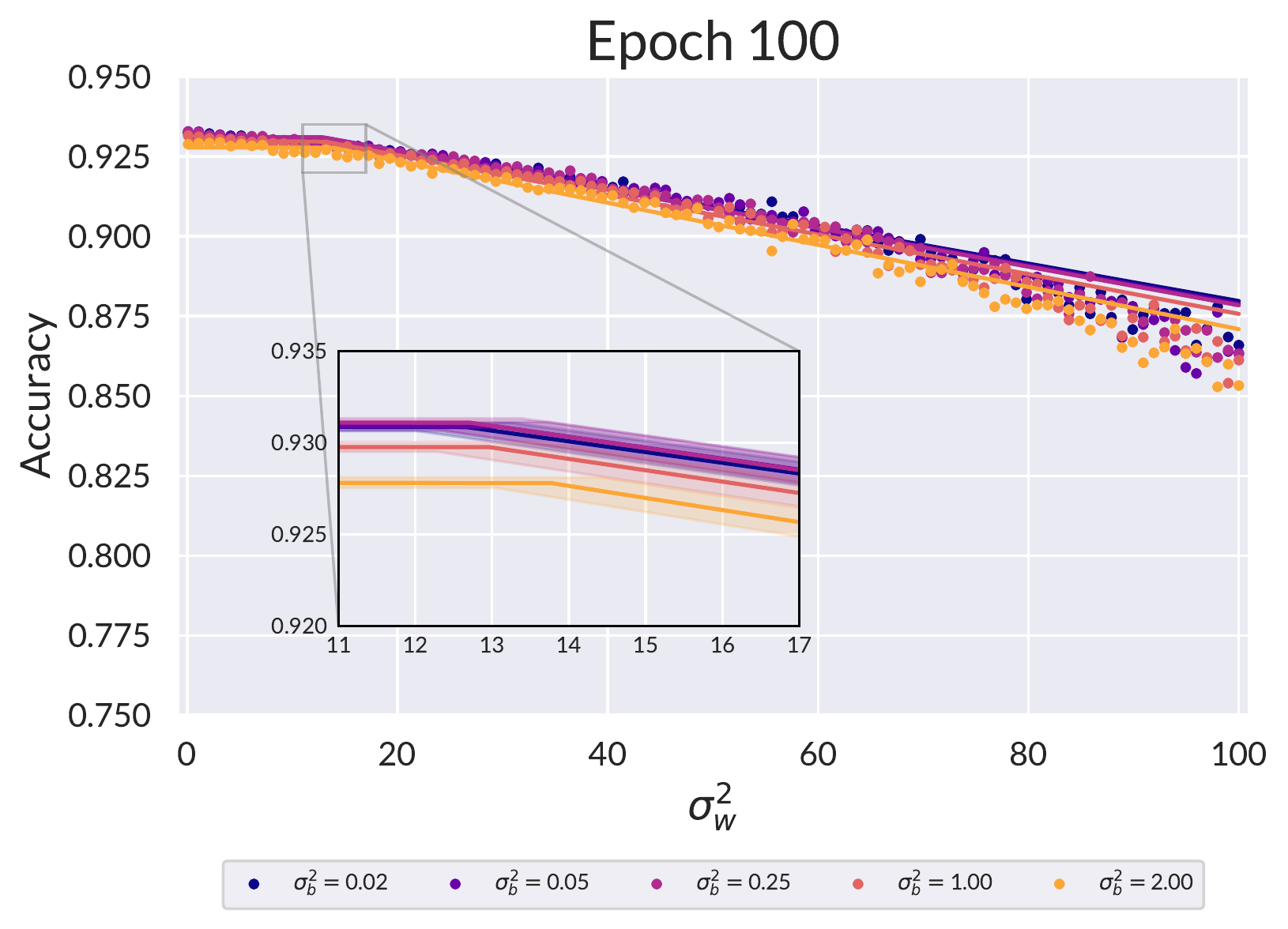}
    \end{subfigure}%
    \begin{subfigure}[b]{9cm}
        \includegraphics[width=8cm]{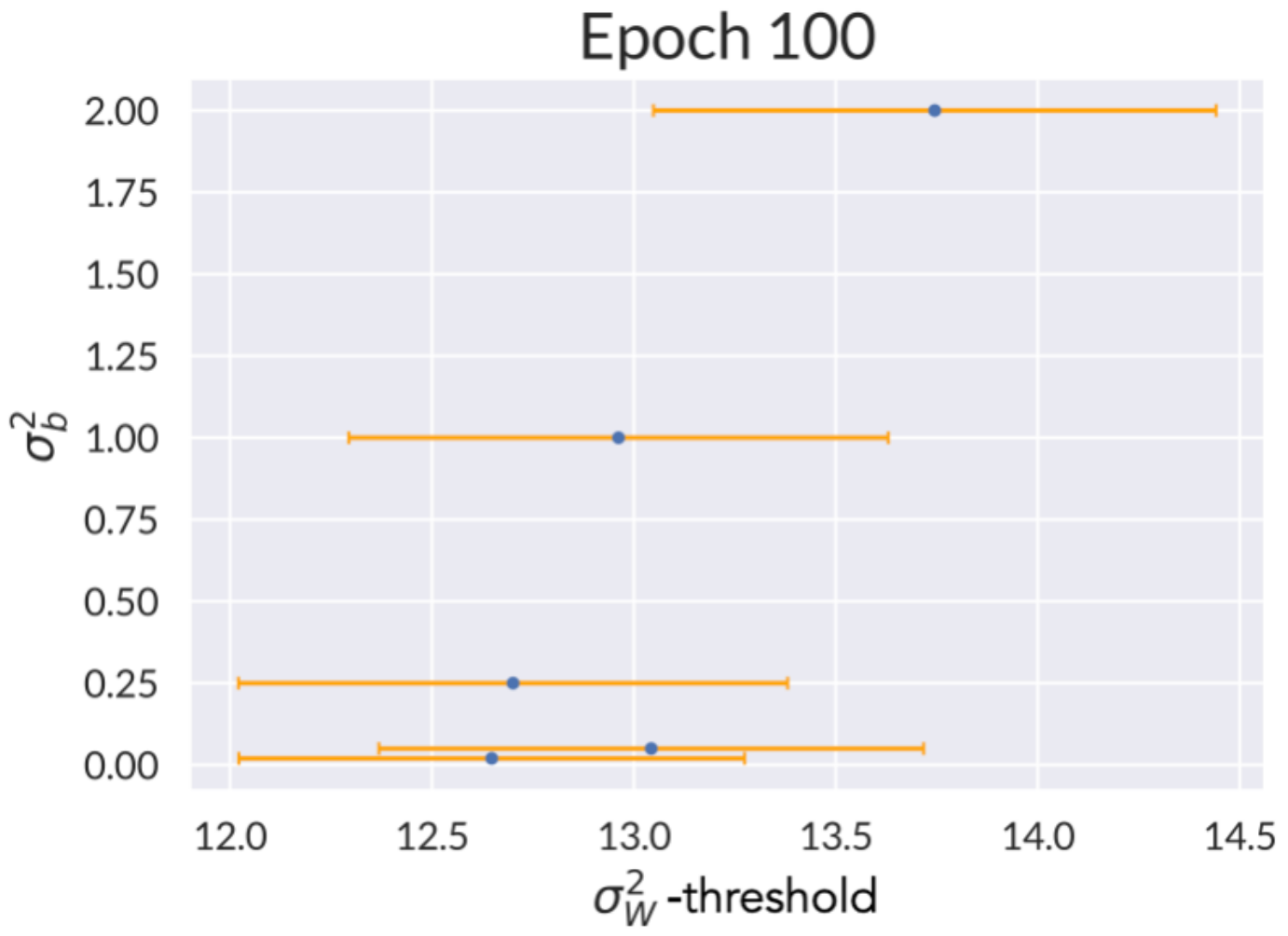}
    \end{subfigure}
    \caption{For small networks, the learning efficiency exhibits threshold behavior as a function of $\sigma_w^2$. Shown are results for MNIST trained on a $N=8$, $L=1$ network sampled over 50 network initializations. The inset shows the fits in the threshold region. The bottom figure shows that the location of this threshold in $\sigma_{w}^{2}$ decreases with increasing $\sigma_b^2$ consistent with the trend implied by the line of uniformity threshold. As explained in the text, there is a multiplicative factor involved and the large-$N$ analysis cannot be straightforwardly transplanted to this small-$N$ case. The uncertainty bars are propagated from the uncertainties in the accuracy versus $\sigma_{w}^{2}$ data points.}
    \label{fig:LOU-threshold}
\end{figure*}

The effect of lying near uniformity is therefore strongest in shallow, narrow networks rather than deep, wide networks where the edge of chaos effect dominates. For these small networks, some of the asymptotic analysis above locating the LOU and EOC does not immediately apply, since the network is unable to reach the asymptotic value $\sigma_{*}^{2}$ of the pre-activation variance.\footnote{In this sense, we may take ``shallow'' to mean $L \leq 5$, since as shown in fig. \ref{fig:postevol}, by $L \approx 6$, the network has reached $\sigma_{*}^{2}$. Strictly speaking however, the predictions for the EOC as well as the LOU are ill-defined in narrow networks, since these are no longer approximately Gaussian, and also appear to be beyond the reach of current perturbative approaches \cite{Grosvenor:2021eol}.} At the same time, the input variance and mean, $\sigma_{0}^{2}$ and $\mu_0$, actually \emph{do} matter in this case and, with this information, we can roughly estimate the location of the line of uniformity. For example, for $L=1$, we have $\sigma_{1}^{2} = \sigma_{w}^{2} ( \sigma_{0}^{2} + \mu_{0}^{2} ) + \sigma_{b}^{2}$ and the line of uniformity would be where $\sigma_{1}^{2} = \sigma_{{\rm min}}^{2} = \frac{\pi^2}{12}$. For example, for MNIST, $\sigma_0^2 \approx 0.095$ and $\mu_{0}^{2} \approx 0.017$, so the line of uniformity can be estimated as $\sigma_b^2 \approx \frac{\pi^2}{12} - 0.112 \sigma_w^2$. Equivalently, for fixed $\sigma_b^2$, this gives a $\sigma_{w}^{2}$-threshold of $\sigma_w^2 \sim 7.35 + 8.93 \sigma_b^2$ beyond which we expect saturation effects to decrease training efficiency. For $L=2$, we would iterate the above process once more, passing through the activation function; this gives an estimated threshold of $\sigma_{w}^{2} \approx 3.5 +8.93\sigma_b^2 $. 

\begin{figure*}
    \centering
    \begin{subfigure}[t]{9cm}
         \includegraphics[width=8cm]{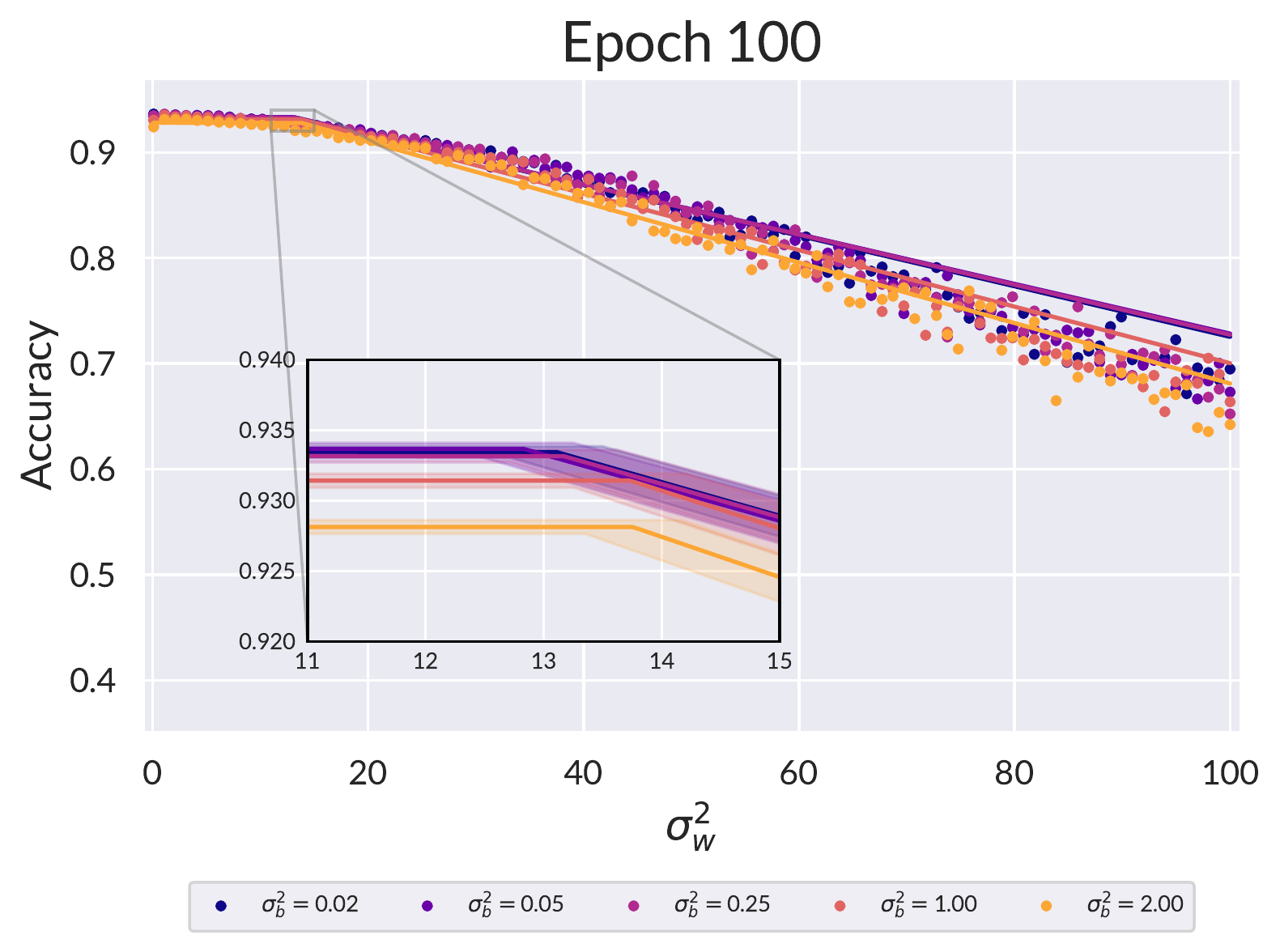}
    \end{subfigure}%
    \begin{subfigure}[b]{9cm}
        \includegraphics[width=8cm]{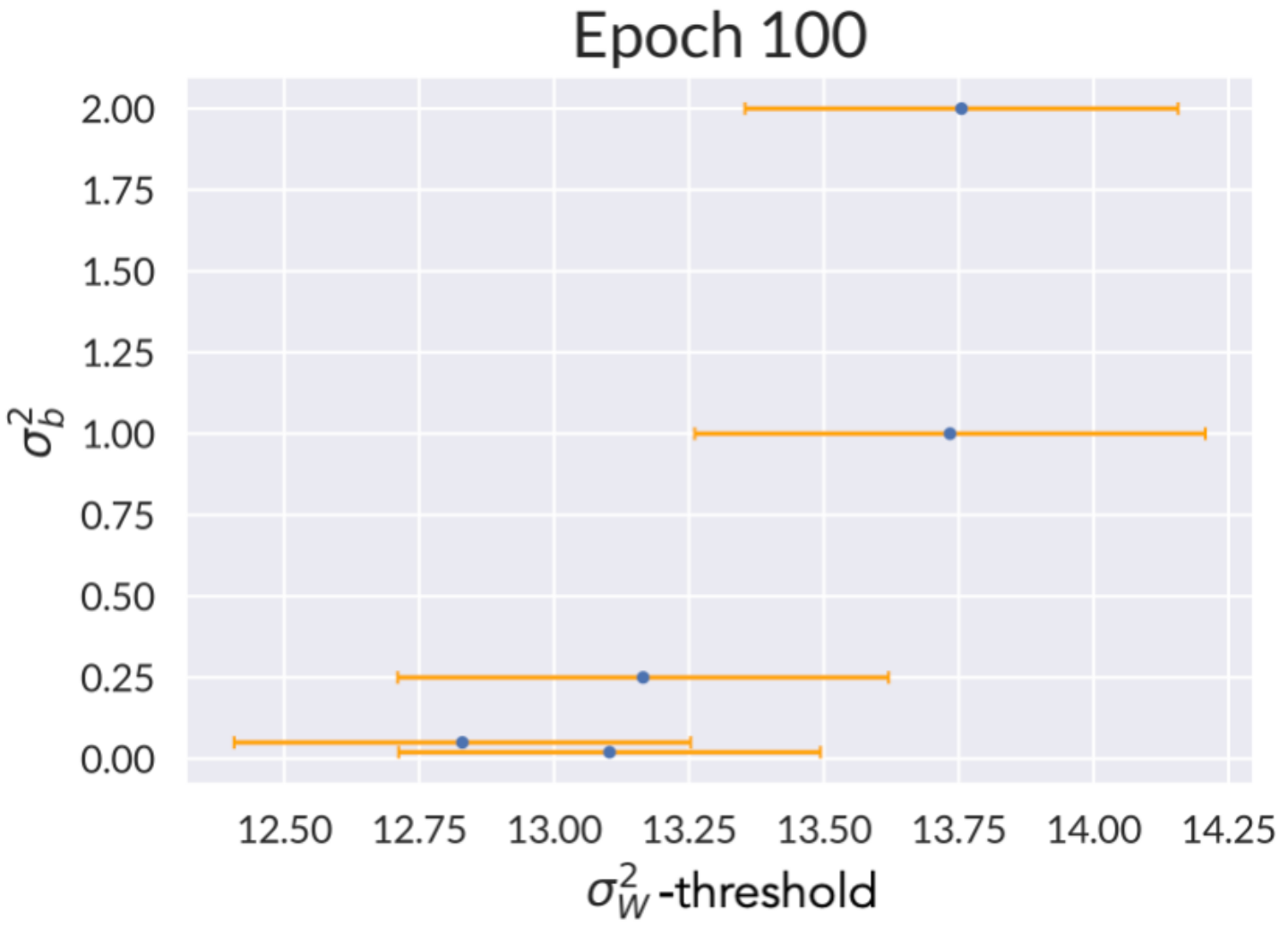}
    \end{subfigure}
\caption{Small-network threshold behavior as in fig. \ref{fig:LOU-threshold}, for MNIST trained on a network with $N=8$ and $L=2$, sampled over 50 initial conditions drawn from $(\sigma_w^2,\sigma_b^2)$. The bottom figure shows that the location of this threshold decreases with $\sigma_b^2$ consistent with the trend implied by the line of uniformity.}
    \label{fig:LOU-threshold_L2}
\end{figure*}

Results for $L=1$ are shown in fig. \ref{fig:LOU-threshold}, and results for $L=2$ are shown in fig. \ref{fig:LOU-threshold_L2}. As predicted, we observe that the accuracy retains a high, approximately constant value up to a $\sigma_b^2$-dependent threshold for $\sigma_w^2$, and then decays approximately linearly thereafter. To determine the threshold empirically, we fit the data to a function of the form
\begin{equation}
    A_{{\rm fit}} ( \sigma_{w}^{2} ) = A_{{\rm max}} - r ( \sigma_{w}^{2} - \sigma_{w, \text{thr}}^{2} ) \, \Theta ( \sigma_{w}^{2} - \sigma_{w, \text{thr}}^{2} ),
\end{equation}
where $A_{{\rm max}}$ is the maximum accuracy, $\sigma_{w, \text{thr}}^{2}$ is the threshold value, $r$ is the rate of linear decay, and $\Theta$ is the Heaviside step function. Each accuracy vs. $\sigma_{w}^{2}$ data point is an average over 20 instantiations of the network and thus comes with its own variance. These propagate into uncertainty bars for the three fit parameters. We plot the threshold for different values of $\sigma_{b}^{2}$ in fig. \ref{fig:LOU-threshold} for $L=1$. This qualitatively confirms our expectations, though the empirical value of the threshold is about a factor of 2 greater than the analytical prediction, and the slope about a factor of 8 smaller. However, given that we are applying a large-$N$ analysis to a relatively narrow network ($N=8$), an $\mathcal{O}(1)$ quantitative discrepancy is reasonable. For $L=2$ the corresponding results are presented in 
fig. \ref{fig:LOU-threshold_L2}, again showing qualitative agreement. The empirical threshold in this case is about a factor of 4 greater than the theoretical value, and the slope is a factor of $8$ smaller.

\emph{Conclusion.} In this work, we establish that for deep random feedforward networks along the edge of chaos, the efficiency of training via stochastic gradient descent still depends on non-saturation of the activation function. Similar points have been made previously in \cite{2018arXiv180508266H,pmlr-v97-hayou19a}, which compared the performance of difference activation functions initialized at one point on their respective edges of chaos. However, what we demonstrate for the tanh activation function is that not all points on the edge of chaos are equally efficient at learning. Within a fixed number of training epochs ($\sim 100$), activation function saturation eventually impedes learning if we push the weight and bias variances too far to the right of the line of uniformity, defined to be where the final layer post-activation is most uniformly distributed, i.e., maximally entropic. Unlike the edge of chaos, which separates chaotic and ordered outputs, the line of uniformity does not mark an abrupt change in the overall behavior of the network. Rather, it simply indicates roughly the point where the saturation of the activation function begins to impede learning. We demonstrate this for shallow and narrow networks as well, where the exponential damping of neuron correlations away from the edge of chaos becomes much less of a decisive factor in determining training efficiency.

\emph{Acknowledgments.} This research was supported 
 in part by the Dutch Research Council
(NWO) project 680-91-116 ({\em Planckian Dissipation and Quantum Thermalisation: From
Black Hole Answers to Strange Metal Questions.}) and
by the Dutch Research
Council (NWO)/Ministry of Education. K.T.G. has received funding from the European Union’s Horizon 2020 research and innovation programme under the Marie Sk\l odowska-Curie grant agreement No 101024967.


\newpage

\onecolumngrid

\appendix


\section{Independence of \texorpdfstring{$\sigma_{*}^{2}$}{sigmastarsquared} on \texorpdfstring{$\sigma_{1}^{2}$}{sigmaonesquared}}
\label{app:postevol}

\begin{figure}[h!]
\centering
\includegraphics[width=\textwidth]{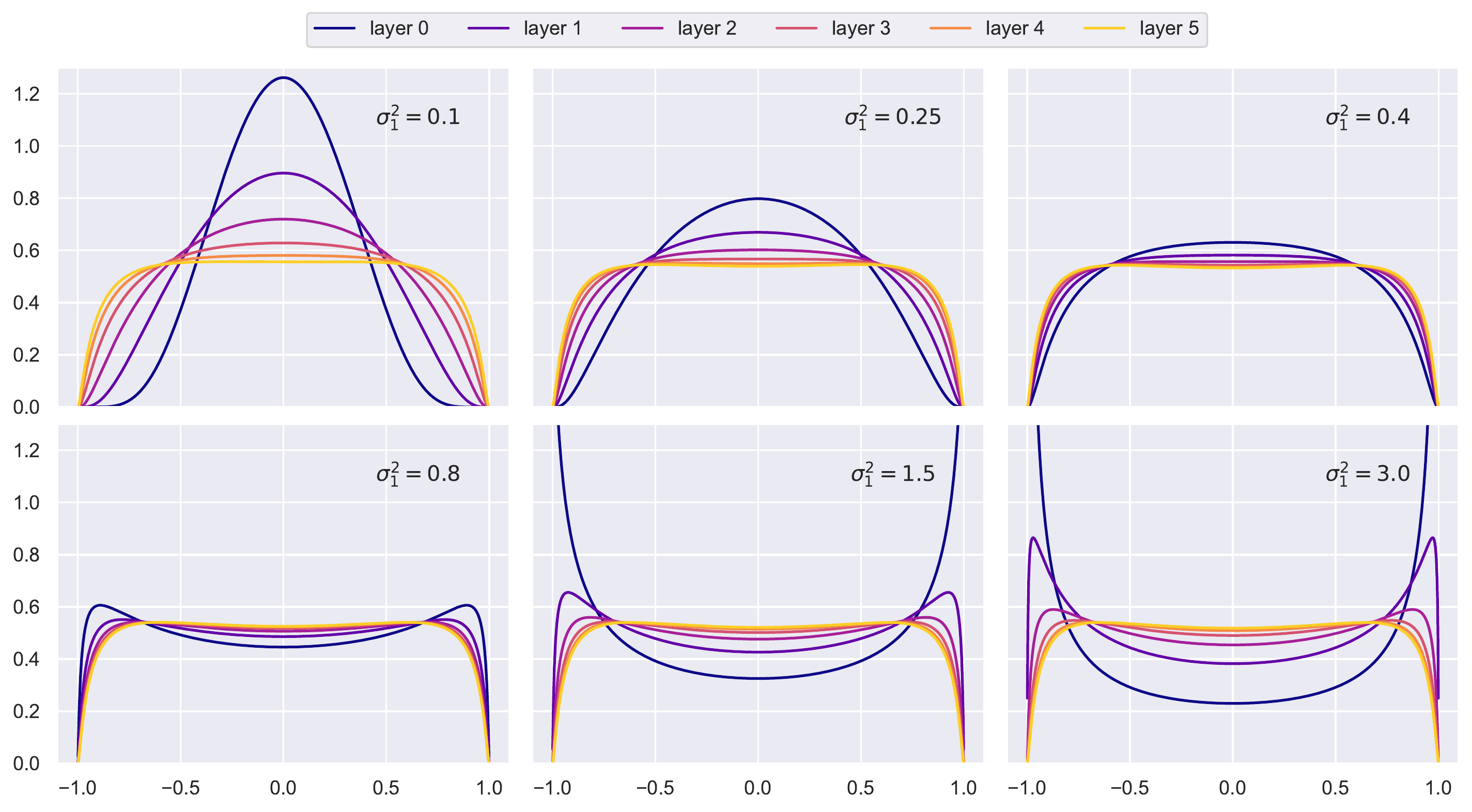}
\caption{Layer-to-layer evolution of the post-activation distribution at $( \sigma_{w}^{2} , \sigma_{b}^{2} ) = (1.76, 0.05)$ for six different values of the first hidden layer pre-activation variance $\sigma_{1}^{2}$. The post-activations converge to the asymptotic distribution within about five layers.}
\label{fig:postevol}
\end{figure}

The exact pre- or post-activation distribution at a given layer obviously does depend on $\sigma_{1}^{2}$, the pre-activation variance at the first hidden layer. This dependence is generated via the recursion relation \eqref{eq:varrecursion}. However, at the fixed point, the asymptotic distributions do not depend on $\sigma_{1}^{2}$. Indeed, the relation that the asymptotic pre-activation variance satisfies is eq. \eqref{eq:sigmastar}, which does not depend on $\sigma_{1}^{2}$ at all. We can demonstrate this fact by plotting the evolution of the post-activation distribution for fixed $\sigma_{w}^{2}$ and $\sigma_{b}^{2}$, but for many values of $\sigma_{1}^{2}$. In fig. \ref{fig:postevol}, we show this for $( \sigma_{w}^{2} , \sigma_{b}^{2} ) = (1.76, 0.05)$ for several values of $\sigma_{1}^{2}$, both less than and greater than $\sigma_{*}^{2}$ which turns out to be $\sigma_{*}^{2} \approx 0.57$ in this case. When $\sigma_{1}^{2} < \sigma_{*}^{2}$, the post-activation distribution starts out narrower and spreads out, whereas when $\sigma_{1}^{2} > \sigma_{*}^{2}$ it starts out more peaked at $\pm 1$ and then flattens out.


\section{Analytic Details of the Fixed Point Computation}
\label{app:analytics}

In this appendix, we will show that the edge of chaos contains the point $( \sigma_{w}^{2} , \sigma_{b}^{2} ) = (1,0)$ and has zero slope there. At this point, the fixed-point equation \eqref{eq:sigmastar} reads
\begin{equation}
    \sigma_{*}^{2} = \sigma_{\phi , *}^{2}~. 
\end{equation}
The left-hand side is the fixed-point pre-activation variance, whereas the right-hand side is the corresponding post-activation variance. As long as $| \phi (z) | < |z|$, which is the case for $\phi (z) = \tanh (z)$ except at $z=0$, the variance of the post-activation will always be smaller than that of the pre-activation. Therefore, the only solution at this point is $\sigma_{*}^{2} = \sigma_{\phi, *}^{2} = 0$ and thus at this point $\phi' ( \sigma_* z ) = \sech^2 (0) = 1$ and $\chi = 1$ or $\xi = \infty$. Hence, this point is on the edge of chaos.

Now, consider eq. \eqref{eq:fixedsigmastar}, but now along the edge of chaos rather than the lines of constant $\sigma_{*}^{2}$. Let $\sigma_{w}^{2}$ be our independent parameter along the edge of chaos and take a derivative with respect to it:
\begin{equation} \label{eq:derrel}
    \frac{\partial \sigma_{b}^{2}}{\partial \sigma_{w}^{2}} = \biggl( 1 - \sigma_{w}^{2} \, \frac{\partial \sigma_{\phi , *}^{2}}{\partial \sigma_{*}^{2}} \biggr) \frac{\partial \sigma_{*}^{2}}{\partial \sigma_{w}^{2}} - \sigma_{\phi , *}^{2}~,
\end{equation}
where we have used the fact that $\sigma_{\phi , *}^{2}$ depends on $\sigma_{w}^{2}$ only through its dependence on $\sigma_{*}^{2}$.

To compute the derivative $\frac{\partial \sigma_{\phi , *}^{2}}{\partial \sigma_{*}^{2}}$, it is convenient to first rewrite the integral expression for $\sigma_{\phi}^{2}$ in \eqref{eq:varpost} by changing back to the original pre-activation variable $z$:
\begin{align}
   \sigma_{\phi}^{2} = \int_{-1}^{1} \mathrm{d} x \, p_{\phi} ( x ; \sigma^2 ) \, x^2 = \int_{- \infty}^{\infty}\mathrm{d} z \, p(z ; \sigma^2 ) \, \phi (z)^2~.
\end{align}
We can easily compute the various derivatives of the pre-activation distribution:
\begin{align}
    \frac{\partial p(z ; \sigma^2 )}{\partial \sigma^2} &= \biggl( \frac{z^2}{\sigma^2} -1 \biggr) \frac{p(z ; \sigma^2 )}{2 \sigma^2}, &%
    %
    %
    \frac{\partial^2 p(z; \sigma^2 )}{\partial z^2} &= \biggl( \frac{z^2}{\sigma^2} -1 \biggr) \frac{p(z ; \sigma^2 )}{\sigma^2} = 2 \, \frac{\partial p(z ; \sigma^2 )}{\partial \sigma^2}~.
\end{align}
Therefore, using integration by parts, and the fact that we can ignore boundary terms due to the fast fall-off of the Gaussian, we find
\begin{align} \label{eq:paderiv}
    \frac{\partial \sigma_{\phi}^{2}}{\partial \sigma^2} = \int \mathrm{d} z \, \frac{\partial p (z ; \sigma^2)}{\partial \sigma^2} \, \phi (z)^2 = \frac{1}{2} \int \mathrm{d} z \, \frac{\partial^2 p(z ; \sigma^2 )}{\partial z^2} \, \phi (z)^2 = \int \mathrm{d} z \, p(z ; \sigma^2 ) \bigl( \phi' (z)^2 + \phi(z) \, \phi'' (z) \bigr)~.
\end{align}
By rescaling the variable to $\sigma z$, the first integral term above can be written as
\begin{equation}
    \int \mathrm{d} z \, p(z; \sigma^2 ) \phi' (z)^2 = \int \mathcal{D} z \, \bigl[ \phi' ( \sigma z ) \bigr]^2~.
\end{equation}
Note that when this is evaluated at $\sigma_{*}^{2}$ and multiplied by $\sigma_{w}^{2}$, we get precisely $\chi$, as defined in \eqref{eq:corlength}. Let us give a name to the remaining integral in \eqref{eq:paderiv} evaluated at $\sigma_{*}^{2}$. For future convenience, we will put a relative minus sign in the definition below, the reason being that, for $\phi = \tanh$, the object $\phi \, \phi''$ is \emph{negative} semi-definite:
\begin{equation}
    \tilde{\chi} = - \sigma_{w}^{2} \int \mathrm{d} z \, p (z ; \sigma_{*}^{2} ) \, \phi (z) \, \phi '' (z) = - \sigma_{w}^{2} \int \mathcal{D} z \, \phi ( \sigma_* z ) \, \phi '' ( \sigma_* z )~.
\end{equation}
Then, \eqref{eq:paderiv} evaluated at $\sigma_{*}^{2}$ and multiplied by $\sigma_{w}^{2}$ reads
\begin{equation} \label{eq:paderiv2}
    \sigma_{2}^{2} \frac{\partial \sigma_{\phi , *}^{2}}{\partial \sigma_{*}^{2}} = \chi - \tilde{\chi}~.
\end{equation}
Now, let us define
\begin{equation}
    \tilde{\xi} = - \frac{1}{\ln ( \chi - \tilde{\chi} )}~.
\end{equation}
This is precisely the object called $\xi_q$ in \cite{2016arXiv161101232S}, which is the length scale that controls the exponential decay of information propagation through the neural network from a single input.

Plugging eq. \eqref{eq:paderiv2} back into eq. \eqref{eq:derrel} gives
\begin{align}
    \frac{\partial \sigma_{b}^{2}}{\partial \sigma_{w}^{2}} = ( 1 - \chi + \tilde{\chi} ) \frac{\partial \sigma_{*}^{2}}{\partial \sigma_{w}^{2}} - \sigma_{\phi , *}^{2}~.
\end{align}
Along the edge of chaos, $\chi = 1$, and so
\begin{align} \label{eq:derrel1}
    \frac{\partial \sigma_{b}^{2}}{\partial \sigma_{w}^{2}} = \tilde{\chi} \frac{\partial \sigma_{*}^{2}}{\partial \sigma_{w}^{2}} - \sigma_{\phi , *}^{2}~,
\end{align}
Now, we can establish a simple bound on $\tilde{\chi}$ by virtue of the fact that $|\phi (z) | \leq |z|$, for $\phi = \tanh$. To do this, let us first rewrite $\tilde{\chi}$ using the identity
\begin{equation}
    \phi'' (z) = - 2 \tanh(z) \sech^2 (z) = -2 \, \phi (z) \, \phi'(z)~.
\end{equation}
Therefore,
\begin{equation}
    \phi (z) \, \phi'' (z) = - 2 \, \phi (z)^2 \, \phi' (z) = - \frac{2}{3} \bigl[ \phi (z)^3 \bigr]'~,
\end{equation}
and
\begin{equation}
    \tilde{\chi} = \frac{2 \sigma_{2}^{2}}{3} \int \mathrm{d} z \, p(z ; \sigma_{*}^{2} ) \bigl[ \phi (z)^3 \bigr]' = - \frac{2 \sigma_{w}^{2}}{3} \int \mathrm{d} z \, \frac{\partial p(z ; \sigma_{*}^{2} )}{\partial z} \, \phi (z)^3 = \frac{2 \sigma_{w}^{2}}{3 \sigma_{*}^{2}} \int \mathrm{d} z \, p (z ; \sigma_{*}^{2} ) \, z \, \phi (z)^3~.
\end{equation}
Therefore, since $|\phi (z) | \leq |z|$ for $\phi = \tanh$,
\begin{align}
    0 \leq \tilde{\chi} \leq \frac{2 \sigma_{w}^{2}}{3 \sigma_{*}^{2}} \int \mathrm{d} z \, p ( z; \sigma_{*}^{2} ) \, z^4 = 2 \, \sigma_{w}^{2} \, \sigma_{*}^{2}~.
\end{align}
Therefore, since we have already shown that $\sigma_{*}^{2} = \sigma_{\phi , *}^{2} = 0$ at the point $( \sigma_{w}^{2} , \sigma_{b}^{2} ) = (1,0)$, it follows that $\tilde{\chi} = 0$ at this point as well and, from eq. \eqref{eq:derrel1},
\begin{equation}
    \frac{\partial \sigma_{b}^{2}}{\partial \sigma_{w}^{2}} \biggr|_{( \sigma_{w}^{2} , \sigma_{b}^{2} ) = (1,0)} = 0~.
\end{equation}
In other words, the edge of chaos has zero slope at the point $( \sigma_{w}^{2} , \sigma_{b}^{2} ) = (1,0)$.


\section{Implementation Details}
\label{app:figs}

\begin{figure}
    \centering
    \includegraphics[width=0.45\textwidth]{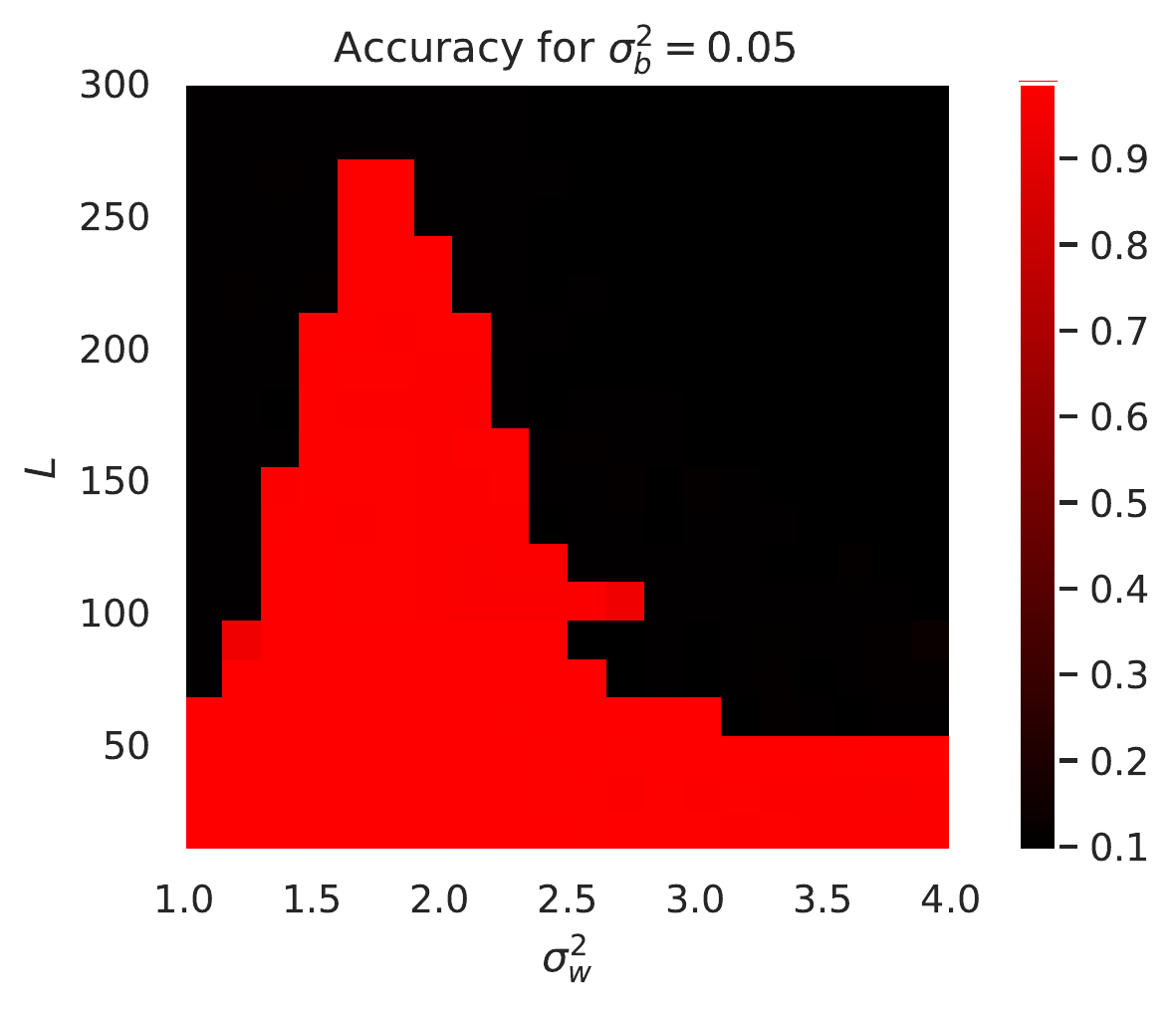}
    \caption{Optimal learning for deep neural networks at the edge of chaos as first shown by \cite{2016arXiv161101232S}. Shown is learning efficiency for MNIST training as a function of network depth $L$ with $N=784$ and choice of initial weight distribution $\sigma_w^2$ holding the initial bias distribution $\sigma_b^2=0.05$ fixed. At the edge of chaos $(\sigma_w^2,\sigma_b^2)=(1.76,0.05)$, learning remains efficient even for very deep networks, but eventually ($L\sim 270$) goes down. This same behavior has been observed for deep feedforward networks in \cite{2016arXiv161101232S,Erdmenger:2021sot}. The learning rate used is $\ell = 10^{-3}$ for $L<100$ and $\ell=10^{-4}$ for $L\geq100$.}
    \label{fig:googleBrain-repro}
\end{figure}

Throughout this work, we have used a vanilla feedforward neural network of $L$ hidden layers, each having the same depth $N$. As described, initial weights and biases are drawn from zero-mean Gaussian distributions with $\frac{\sigma_w^2}{N}$ and $\sigma_b^2$ respectively. Both MNIST and CIFAR-10 were trained using the standard cross-entropy loss function and no optimizer. This reproduces the results of \cite{2016arXiv161101232S} (see fig. \ref{fig:googleBrain-repro}), confirming critical behavior.

\section{SWISH activation function}\label{app:swish}

\begin{figure*}[h!]
\centering
\begin{subfigure}{8.9cm}
\centering
\includegraphics[width=8.2cm]{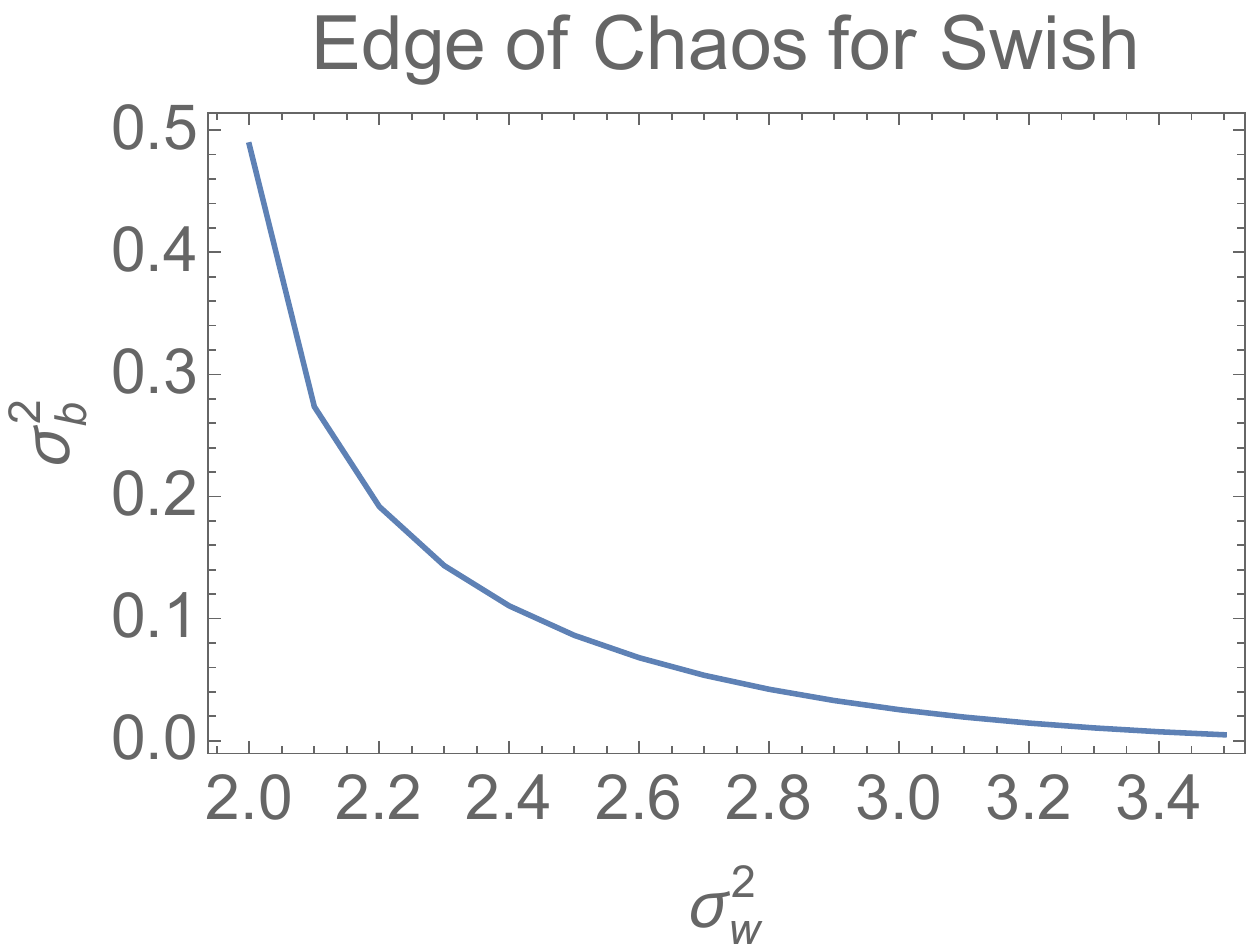}
\end{subfigure}
\begin{subfigure}{8.9cm}
\centering
\includegraphics[width=8.9cm]{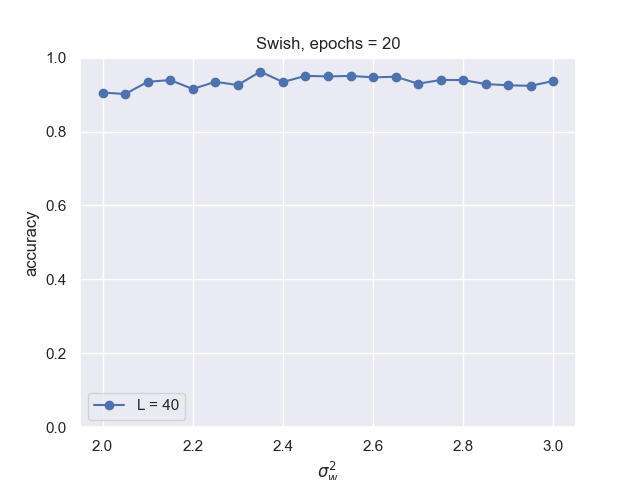}
\end{subfigure}
\caption{(Left) Edge of chaos for SWISH activation function. (Right) Accuracy for a feedforward network with $L=40$ layers trained on MNIST for 21 equally-spaced points along the SWISH EOC. Over the limited range for which the criticality condition $\chi=1$ is satisfied, we observe no significant differences in accuracy, though a slightly lower learning rate was used for the left-most two points; we believe this to be due to the large values of $\sigma_b^2$ in this regime. \label{fig:swisheoc}}
\end{figure*}

Throughout the text, we examined the impact of saturation via the line of uniformity for the tanh activation function. For non-saturating activation functions, it is an open question whether a similar notion of uniformity exists. While a full analysis of this is beyond the scope of this work, in this appendix we offer some preliminary results for the SWISH activation function,
\begin{equation}
	\mathrm{swish}(z)=\frac{z}{1+e^{-z}}~,
\end{equation}
which also features a line of critical points separating an ordered and chaotic phase. Note that unlike the EOC for tanh, which increases with increasing $\sigma_w^2$, the EOC for SWISH decreases with increasing $\sigma_w^2$, which prevents us from examining the impact of large weight variances. Conversely, for small values of $\sigma_w^2$, the corresponding value of $\sigma_b^2$ becomes so large that we are unable to satisfy the critical detection criteria $\chi=1$ discussed in the main text.\footnote{We do not claim that the EOC stops beyond this point, rather that it cannot be computed from the central limit method used in \cite{arxiv.1606.05340, 2016arXiv161101232S}. It is conceivable that this could be computed via the NN/QFT correspondence developed in \cite{Grosvenor:2021eol}, but this has not been attempted for SWISH.} The EOC for SWISH is plotted in fig. \ref{fig:swisheoc}, which shows a computable range of approximately $\sigma_w^2\in[1.97,3.4]$. The same figure also shows the accuracy for an $L=40$ network with SWISH activation function trained along the EOC, demonstrating no deterioration of performance within this range, which confirms the absence of saturation effects. See also \cite{2018arXiv180508266H,pmlr-v97-hayou19a}.

\bibliographystyle{apsrev4-1}
\bibliography{eoc_uniformity}

\begin{thebibliography}{19}%
\makeatletter
\providecommand \@ifxundefined [1]{%
 \@ifx{#1\undefined}
}%
\providecommand \@ifnum [1]{%
 \ifnum #1\expandafter \@firstoftwo
 \else \expandafter \@secondoftwo
 \fi
}%
\providecommand \@ifx [1]{%
 \ifx #1\expandafter \@firstoftwo
 \else \expandafter \@secondoftwo
 \fi
}%
\providecommand \natexlab [1]{#1}%
\providecommand \enquote  [1]{``#1''}%
\providecommand \bibnamefont  [1]{#1}%
\providecommand \bibfnamefont [1]{#1}%
\providecommand \citenamefont [1]{#1}%
\providecommand \href@noop [0]{\@secondoftwo}%
\providecommand \href [0]{\begingroup \@sanitize@url \@href}%
\providecommand \@href[1]{\@@startlink{#1}\@@href}%
\providecommand \@@href[1]{\endgroup#1\@@endlink}%
\providecommand \@sanitize@url [0]{\catcode `\\12\catcode `\$12\catcode
  `\&12\catcode `\#12\catcode `\^12\catcode `\_12\catcode `\%12\relax}%
\providecommand \@@startlink[1]{}%
\providecommand \@@endlink[0]{}%
\providecommand \url  [0]{\begingroup\@sanitize@url \@url }%
\providecommand \@url [1]{\endgroup\@href {#1}{\urlprefix }}%
\providecommand \urlprefix  [0]{URL }%
\providecommand \Eprint [0]{\href }%
\providecommand \doibase [0]{http://dx.doi.org/}%
\providecommand \selectlanguage [0]{\@gobble}%
\providecommand \bibinfo  [0]{\@secondoftwo}%
\providecommand \bibfield  [0]{\@secondoftwo}%
\providecommand \translation [1]{[#1]}%
\providecommand \BibitemOpen [0]{}%
\providecommand \bibitemStop [0]{}%
\providecommand \bibitemNoStop [0]{.\EOS\space}%
\providecommand \EOS [0]{\spacefactor3000\relax}%
\providecommand \BibitemShut  [1]{\csname bibitem#1\endcsname}%
\let\auto@bib@innerbib\@empty
\bibitem [{\citenamefont {Krizhevsky}\ \emph {et~al.}(2012)\citenamefont
  {Krizhevsky}, \citenamefont {Sutskever},\ and\ \citenamefont
  {Hinton}}]{NIPS2012_c399862d}%
  \BibitemOpen
  \bibfield  {author} {\bibinfo {author} {\bibfnamefont {A.}~\bibnamefont
  {Krizhevsky}}, \bibinfo {author} {\bibfnamefont {I.}~\bibnamefont
  {Sutskever}}, \ and\ \bibinfo {author} {\bibfnamefont {G.~E.}\ \bibnamefont
  {Hinton}},\ }\bibfield  {title} {\enquote {\bibinfo {title} {Imagenet
  classification with deep convolutional neural networks},}\ }in\ \href
  {https://proceedings.neurips.cc/paper/2012/file/c399862d3b9d6b76c8436e924a68c45b-Paper.pdf}
  {\emph {\bibinfo {booktitle} {Advances in Neural Information Processing
  Systems}}},\ Vol.~\bibinfo {volume} {25},\ \bibinfo {editor} {edited by\
  \bibinfo {editor} {\bibfnamefont {F.}~\bibnamefont {Pereira}}, \bibinfo
  {editor} {\bibfnamefont {C.}~\bibnamefont {Burges}}, \bibinfo {editor}
  {\bibfnamefont {L.}~\bibnamefont {Bottou}}, \ and\ \bibinfo {editor}
  {\bibfnamefont {K.}~\bibnamefont {Weinberger}}}\ (\bibinfo  {publisher}
  {Curran Associates, Inc.},\ \bibinfo {year} {2012})\BibitemShut {NoStop}%
\bibitem [{\citenamefont {{Ramesh}}\ \emph {et~al.}(2021)\citenamefont
  {{Ramesh}}, \citenamefont {{Pavlov}}, \citenamefont {{Goh}}, \citenamefont
  {{Gray}}, \citenamefont {{Voss}}, \citenamefont {{Radford}}, \citenamefont
  {{Chen}},\ and\ \citenamefont {{Sutskever}}}]{2021arXiv210212092R}%
  \BibitemOpen
  \bibfield  {author} {\bibinfo {author} {\bibfnamefont {A.}~\bibnamefont
  {{Ramesh}}}, \bibinfo {author} {\bibfnamefont {M.}~\bibnamefont {{Pavlov}}},
  \bibinfo {author} {\bibfnamefont {G.}~\bibnamefont {{Goh}}}, \bibinfo
  {author} {\bibfnamefont {S.}~\bibnamefont {{Gray}}}, \bibinfo {author}
  {\bibfnamefont {C.}~\bibnamefont {{Voss}}}, \bibinfo {author} {\bibfnamefont
  {A.}~\bibnamefont {{Radford}}}, \bibinfo {author} {\bibfnamefont
  {M.}~\bibnamefont {{Chen}}}, \ and\ \bibinfo {author} {\bibfnamefont
  {I.}~\bibnamefont {{Sutskever}}},\ }\bibfield  {title} {\enquote {\bibinfo
  {title} {{Zero-Shot Text-to-Image Generation}},}\ }\href@noop {} {\bibfield
  {journal} {\bibinfo  {journal} {arXiv e-prints}\ ,\ \bibinfo {eid}
  {arXiv:2102.12092}} (\bibinfo {year} {2021})},\ \Eprint
  {http://arxiv.org/abs/2102.12092} {arXiv:2102.12092 [cs.CV]} \BibitemShut
  {NoStop}%
\bibitem [{\citenamefont {{Brown}}\ \emph {et~al.}(2020)\citenamefont
  {{Brown}}, \citenamefont {{Mann}}, \citenamefont {{Ryder}}, \citenamefont
  {{Subbiah}}, \citenamefont {{Kaplan}}, \citenamefont {{Dhariwal}},
  \citenamefont {{Neelakantan}}, \citenamefont {{Shyam}}, \citenamefont
  {{Sastry}}, \citenamefont {{Askell}}, \citenamefont {{Agarwal}},
  \citenamefont {{Herbert-Voss}}, \citenamefont {{Krueger}}, \citenamefont
  {{Henighan}}, \citenamefont {{Child}}, \citenamefont {{Ramesh}},
  \citenamefont {{Ziegler}}, \citenamefont {{Wu}}, \citenamefont {{Winter}},
  \citenamefont {{Hesse}}, \citenamefont {{Chen}}, \citenamefont {{Sigler}},
  \citenamefont {{Litwin}}, \citenamefont {{Gray}}, \citenamefont {{Chess}},
  \citenamefont {{Clark}}, \citenamefont {{Berner}}, \citenamefont
  {{McCandlish}}, \citenamefont {{Radford}}, \citenamefont {{Sutskever}},\ and\
  \citenamefont {{Amodei}}}]{2020arXiv200514165B}%
  \BibitemOpen
  \bibfield  {author} {\bibinfo {author} {\bibfnamefont {T.~B.}\ \bibnamefont
  {{Brown}}}, \bibinfo {author} {\bibfnamefont {B.}~\bibnamefont {{Mann}}},
  \bibinfo {author} {\bibfnamefont {N.}~\bibnamefont {{Ryder}}}, \bibinfo
  {author} {\bibfnamefont {M.}~\bibnamefont {{Subbiah}}}, \bibinfo {author}
  {\bibfnamefont {J.}~\bibnamefont {{Kaplan}}}, \bibinfo {author}
  {\bibfnamefont {P.}~\bibnamefont {{Dhariwal}}}, \bibinfo {author}
  {\bibfnamefont {A.}~\bibnamefont {{Neelakantan}}}, \bibinfo {author}
  {\bibfnamefont {P.}~\bibnamefont {{Shyam}}}, \bibinfo {author} {\bibfnamefont
  {G.}~\bibnamefont {{Sastry}}}, \bibinfo {author} {\bibfnamefont
  {A.}~\bibnamefont {{Askell}}}, \bibinfo {author} {\bibfnamefont
  {S.}~\bibnamefont {{Agarwal}}}, \bibinfo {author} {\bibfnamefont
  {A.}~\bibnamefont {{Herbert-Voss}}}, \bibinfo {author} {\bibfnamefont
  {G.}~\bibnamefont {{Krueger}}}, \bibinfo {author} {\bibfnamefont
  {T.}~\bibnamefont {{Henighan}}}, \bibinfo {author} {\bibfnamefont
  {R.}~\bibnamefont {{Child}}}, \bibinfo {author} {\bibfnamefont
  {A.}~\bibnamefont {{Ramesh}}}, \bibinfo {author} {\bibfnamefont {D.~M.}\
  \bibnamefont {{Ziegler}}}, \bibinfo {author} {\bibfnamefont {J.}~\bibnamefont
  {{Wu}}}, \bibinfo {author} {\bibfnamefont {C.}~\bibnamefont {{Winter}}},
  \bibinfo {author} {\bibfnamefont {C.}~\bibnamefont {{Hesse}}}, \bibinfo
  {author} {\bibfnamefont {M.}~\bibnamefont {{Chen}}}, \bibinfo {author}
  {\bibfnamefont {E.}~\bibnamefont {{Sigler}}}, \bibinfo {author}
  {\bibfnamefont {M.}~\bibnamefont {{Litwin}}}, \bibinfo {author}
  {\bibfnamefont {S.}~\bibnamefont {{Gray}}}, \bibinfo {author} {\bibfnamefont
  {B.}~\bibnamefont {{Chess}}}, \bibinfo {author} {\bibfnamefont
  {J.}~\bibnamefont {{Clark}}}, \bibinfo {author} {\bibfnamefont
  {C.}~\bibnamefont {{Berner}}}, \bibinfo {author} {\bibfnamefont
  {S.}~\bibnamefont {{McCandlish}}}, \bibinfo {author} {\bibfnamefont
  {A.}~\bibnamefont {{Radford}}}, \bibinfo {author} {\bibfnamefont
  {I.}~\bibnamefont {{Sutskever}}}, \ and\ \bibinfo {author} {\bibfnamefont
  {D.}~\bibnamefont {{Amodei}}},\ }\bibfield  {title} {\enquote {\bibinfo
  {title} {{Language Models are Few-Shot Learners}},}\ }\href@noop {}
  {\bibfield  {journal} {\bibinfo  {journal} {arXiv e-prints}\ ,\ \bibinfo
  {eid} {arXiv:2005.14165}} (\bibinfo {year} {2020})},\ \Eprint
  {http://arxiv.org/abs/2005.14165} {arXiv:2005.14165 [cs.CL]} \BibitemShut
  {NoStop}%
\bibitem [{\citenamefont {Jumper}\ \emph {et~al.}(2021)\citenamefont {Jumper},
  \citenamefont {Evans}, \citenamefont {Pritzel}, \citenamefont {Green},
  \citenamefont {Figurnov}, \citenamefont {Ronneberger}, \citenamefont
  {Tunyasuvunakool}, \citenamefont {Bates}, \citenamefont {{\v Z}{\'\i}dek},
  \citenamefont {Potapenko}, \citenamefont {Bridgland}, \citenamefont {Meyer},
  \citenamefont {Kohl}, \citenamefont {Ballard}, \citenamefont {Cowie},
  \citenamefont {Romera-Paredes}, \citenamefont {Nikolov}, \citenamefont
  {Jain}, \citenamefont {Adler}, \citenamefont {Back}, \citenamefont
  {Petersen}, \citenamefont {Reiman}, \citenamefont {Clancy}, \citenamefont
  {Zielinski}, \citenamefont {Steinegger}, \citenamefont {Pacholska},
  \citenamefont {Berghammer}, \citenamefont {Bodenstein}, \citenamefont
  {Silver}, \citenamefont {Vinyals}, \citenamefont {Senior}, \citenamefont
  {Kavukcuoglu}, \citenamefont {Kohli},\ and\ \citenamefont
  {Hassabis}}]{Jumper:2021wp}%
  \BibitemOpen
  \bibfield  {author} {\bibinfo {author} {\bibfnamefont {J.}~\bibnamefont
  {Jumper}}, \bibinfo {author} {\bibfnamefont {R.}~\bibnamefont {Evans}},
  \bibinfo {author} {\bibfnamefont {A.}~\bibnamefont {Pritzel}}, \bibinfo
  {author} {\bibfnamefont {T.}~\bibnamefont {Green}}, \bibinfo {author}
  {\bibfnamefont {M.}~\bibnamefont {Figurnov}}, \bibinfo {author}
  {\bibfnamefont {O.}~\bibnamefont {Ronneberger}}, \bibinfo {author}
  {\bibfnamefont {K.}~\bibnamefont {Tunyasuvunakool}}, \bibinfo {author}
  {\bibfnamefont {R.}~\bibnamefont {Bates}}, \bibinfo {author} {\bibfnamefont
  {A.}~\bibnamefont {{\v Z}{\'\i}dek}}, \bibinfo {author} {\bibfnamefont
  {A.}~\bibnamefont {Potapenko}}, \bibinfo {author} {\bibfnamefont
  {A.}~\bibnamefont {Bridgland}}, \bibinfo {author} {\bibfnamefont
  {C.}~\bibnamefont {Meyer}}, \bibinfo {author} {\bibfnamefont {S.~A.~A.}\
  \bibnamefont {Kohl}}, \bibinfo {author} {\bibfnamefont {A.~J.}\ \bibnamefont
  {Ballard}}, \bibinfo {author} {\bibfnamefont {A.}~\bibnamefont {Cowie}},
  \bibinfo {author} {\bibfnamefont {B.}~\bibnamefont {Romera-Paredes}},
  \bibinfo {author} {\bibfnamefont {S.}~\bibnamefont {Nikolov}}, \bibinfo
  {author} {\bibfnamefont {R.}~\bibnamefont {Jain}}, \bibinfo {author}
  {\bibfnamefont {J.}~\bibnamefont {Adler}}, \bibinfo {author} {\bibfnamefont
  {T.}~\bibnamefont {Back}}, \bibinfo {author} {\bibfnamefont {S.}~\bibnamefont
  {Petersen}}, \bibinfo {author} {\bibfnamefont {D.}~\bibnamefont {Reiman}},
  \bibinfo {author} {\bibfnamefont {E.}~\bibnamefont {Clancy}}, \bibinfo
  {author} {\bibfnamefont {M.}~\bibnamefont {Zielinski}}, \bibinfo {author}
  {\bibfnamefont {M.}~\bibnamefont {Steinegger}}, \bibinfo {author}
  {\bibfnamefont {M.}~\bibnamefont {Pacholska}}, \bibinfo {author}
  {\bibfnamefont {T.}~\bibnamefont {Berghammer}}, \bibinfo {author}
  {\bibfnamefont {S.}~\bibnamefont {Bodenstein}}, \bibinfo {author}
  {\bibfnamefont {D.}~\bibnamefont {Silver}}, \bibinfo {author} {\bibfnamefont
  {O.}~\bibnamefont {Vinyals}}, \bibinfo {author} {\bibfnamefont {A.~W.}\
  \bibnamefont {Senior}}, \bibinfo {author} {\bibfnamefont {K.}~\bibnamefont
  {Kavukcuoglu}}, \bibinfo {author} {\bibfnamefont {P.}~\bibnamefont {Kohli}},
  \ and\ \bibinfo {author} {\bibfnamefont {D.}~\bibnamefont {Hassabis}},\
  }\bibfield  {title} {\enquote {\bibinfo {title} {Highly accurate protein
  structure prediction with alphafold},}\ }\href {\doibase
  10.1038/s41586-021-03819-2} {\bibfield  {journal} {\bibinfo  {journal}
  {Nature}\ }\textbf {\bibinfo {volume} {596}},\ \bibinfo {pages} {583--589}
  (\bibinfo {year} {2021})}\BibitemShut {NoStop}%
\bibitem [{\citenamefont {Schrittwieser}\ \emph {et~al.}(2020)\citenamefont
  {Schrittwieser}, \citenamefont {Antonoglou}, \citenamefont {Hubert},
  \citenamefont {Simonyan}, \citenamefont {Sifre}, \citenamefont {Schmitt},
  \citenamefont {Guez}, \citenamefont {Lockhart}, \citenamefont {Hassabis},
  \citenamefont {Graepel}, \citenamefont {Lillicrap},\ and\ \citenamefont
  {Silver}}]{Schrittwieser:2020ti}%
  \BibitemOpen
  \bibfield  {author} {\bibinfo {author} {\bibfnamefont {J.}~\bibnamefont
  {Schrittwieser}}, \bibinfo {author} {\bibfnamefont {I.}~\bibnamefont
  {Antonoglou}}, \bibinfo {author} {\bibfnamefont {T.}~\bibnamefont {Hubert}},
  \bibinfo {author} {\bibfnamefont {K.}~\bibnamefont {Simonyan}}, \bibinfo
  {author} {\bibfnamefont {L.}~\bibnamefont {Sifre}}, \bibinfo {author}
  {\bibfnamefont {S.}~\bibnamefont {Schmitt}}, \bibinfo {author} {\bibfnamefont
  {A.}~\bibnamefont {Guez}}, \bibinfo {author} {\bibfnamefont {E.}~\bibnamefont
  {Lockhart}}, \bibinfo {author} {\bibfnamefont {D.}~\bibnamefont {Hassabis}},
  \bibinfo {author} {\bibfnamefont {T.}~\bibnamefont {Graepel}}, \bibinfo
  {author} {\bibfnamefont {T.}~\bibnamefont {Lillicrap}}, \ and\ \bibinfo
  {author} {\bibfnamefont {D.}~\bibnamefont {Silver}},\ }\bibfield  {title}
  {\enquote {\bibinfo {title} {Mastering atari, go, chess and shogi by planning
  with a learned model},}\ }\href {\doibase 10.1038/s41586-020-03051-4}
  {\bibfield  {journal} {\bibinfo  {journal} {Nature}\ }\textbf {\bibinfo
  {volume} {588}},\ \bibinfo {pages} {604--609} (\bibinfo {year}
  {2020})}\BibitemShut {NoStop}%
\bibitem [{\citenamefont {{Raghu}}\ \emph {et~al.}(2016)\citenamefont
  {{Raghu}}, \citenamefont {{Poole}}, \citenamefont {{Kleinberg}},
  \citenamefont {{Ganguli}},\ and\ \citenamefont
  {{Sohl-Dickstein}}}]{2016arXiv160605336R}%
  \BibitemOpen
  \bibfield  {author} {\bibinfo {author} {\bibfnamefont {M.}~\bibnamefont
  {{Raghu}}}, \bibinfo {author} {\bibfnamefont {B.}~\bibnamefont {{Poole}}},
  \bibinfo {author} {\bibfnamefont {J.}~\bibnamefont {{Kleinberg}}}, \bibinfo
  {author} {\bibfnamefont {S.}~\bibnamefont {{Ganguli}}}, \ and\ \bibinfo
  {author} {\bibfnamefont {J.}~\bibnamefont {{Sohl-Dickstein}}},\ }\bibfield
  {title} {\enquote {\bibinfo {title} {{On the Expressive Power of Deep Neural
  Networks}},}\ }\href@noop {} {\bibfield  {journal} {\bibinfo  {journal}
  {arXiv e-prints}\ ,\ \bibinfo {eid} {arXiv:1606.05336}} (\bibinfo {year}
  {2016})},\ \Eprint {http://arxiv.org/abs/1606.05336} {arXiv:1606.05336
  [stat.ML]} \BibitemShut {NoStop}%
\bibitem [{\citenamefont {G{\'e}ron}(2019)}]{geron2019hands}%
  \BibitemOpen
  \bibfield  {author} {\bibinfo {author} {\bibfnamefont {A.}~\bibnamefont
  {G{\'e}ron}},\ }\href@noop {} {\emph {\bibinfo {title} {Hands-on Machine
  Learning with Scikit-Learn, Keras, and TensorFlow: Unsupervised learning
  techniques}}}\ (\bibinfo  {publisher} {O'Reilly Media, Incorporated},\
  \bibinfo {year} {2019})\BibitemShut {NoStop}%
\bibitem [{\citenamefont {Poole}\ \emph {et~al.}(2016)\citenamefont {Poole},
  \citenamefont {Lahiri}, \citenamefont {Raghu}, \citenamefont
  {Sohl-Dickstein},\ and\ \citenamefont {Ganguli}}]{arxiv.1606.05340}%
  \BibitemOpen
  \bibfield  {author} {\bibinfo {author} {\bibfnamefont {B.}~\bibnamefont
  {Poole}}, \bibinfo {author} {\bibfnamefont {S.}~\bibnamefont {Lahiri}},
  \bibinfo {author} {\bibfnamefont {M.}~\bibnamefont {Raghu}}, \bibinfo
  {author} {\bibfnamefont {J.}~\bibnamefont {Sohl-Dickstein}}, \ and\ \bibinfo
  {author} {\bibfnamefont {S.}~\bibnamefont {Ganguli}},\ }\bibfield  {title}
  {\enquote {\bibinfo {title} {Exponential expressivity in deep neural networks
  through transient chaos},}\ }\href {\doibase 10.48550/arXiv.1606.05340} {\
  (\bibinfo {year} {2016}),\ 10.48550/arXiv.1606.05340}\BibitemShut {NoStop}%
\bibitem [{\citenamefont {{Schoenholz}}\ \emph {et~al.}(2016)\citenamefont
  {{Schoenholz}}, \citenamefont {{Gilmer}}, \citenamefont {{Ganguli}},\ and\
  \citenamefont {{Sohl-Dickstein}}}]{2016arXiv161101232S}%
  \BibitemOpen
  \bibfield  {author} {\bibinfo {author} {\bibfnamefont {S.~S.}\ \bibnamefont
  {{Schoenholz}}}, \bibinfo {author} {\bibfnamefont {J.}~\bibnamefont
  {{Gilmer}}}, \bibinfo {author} {\bibfnamefont {S.}~\bibnamefont {{Ganguli}}},
  \ and\ \bibinfo {author} {\bibfnamefont {J.}~\bibnamefont
  {{Sohl-Dickstein}}},\ }\bibfield  {title} {\enquote {\bibinfo {title} {{Deep
  Information Propagation}},}\ }\href@noop {} {\bibfield  {journal} {\bibinfo
  {journal} {arXiv e-prints}\ ,\ \bibinfo {eid} {arXiv:1611.01232}} (\bibinfo
  {year} {2016})},\ \Eprint {http://arxiv.org/abs/1611.01232} {arXiv:1611.01232
  [stat.ML]} \BibitemShut {NoStop}%
\bibitem [{\citenamefont {Xiao}\ \emph {et~al.}(2018)\citenamefont {Xiao},
  \citenamefont {Bahri}, \citenamefont {Sohl-Dickstein}, \citenamefont
  {Schoenholz},\ and\ \citenamefont {Pennington}}]{arxiv.1806.05393}%
  \BibitemOpen
  \bibfield  {author} {\bibinfo {author} {\bibfnamefont {L.}~\bibnamefont
  {Xiao}}, \bibinfo {author} {\bibfnamefont {Y.}~\bibnamefont {Bahri}},
  \bibinfo {author} {\bibfnamefont {J.}~\bibnamefont {Sohl-Dickstein}},
  \bibinfo {author} {\bibfnamefont {S.~S.}\ \bibnamefont {Schoenholz}}, \ and\
  \bibinfo {author} {\bibfnamefont {J.}~\bibnamefont {Pennington}},\ }\bibfield
   {title} {\enquote {\bibinfo {title} {Dynamical isometry and a mean field
  theory of cnns: How to train 10,000-layer vanilla convolutional neural
  networks},}\ }\href {\doibase 10.48550/arXiv.1806.05393} {\  (\bibinfo {year}
  {2018}),\ 10.48550/arXiv.1806.05393}\BibitemShut {NoStop}%
\bibitem [{\citenamefont {Chen}\ \emph {et~al.}(2018)\citenamefont {Chen},
  \citenamefont {Pennington},\ and\ \citenamefont
  {Schoenholz}}]{arxiv.1806.05394}%
  \BibitemOpen
  \bibfield  {author} {\bibinfo {author} {\bibfnamefont {M.}~\bibnamefont
  {Chen}}, \bibinfo {author} {\bibfnamefont {J.}~\bibnamefont {Pennington}}, \
  and\ \bibinfo {author} {\bibfnamefont {S.~S.}\ \bibnamefont {Schoenholz}},\
  }\bibfield  {title} {\enquote {\bibinfo {title} {Dynamical isometry and a
  mean field theory of rnns: Gating enables signal propagation in recurrent
  neural networks},}\ }\href {\doibase 10.48550/arXiv.1806.05394} {\  (\bibinfo
  {year} {2018}),\ 10.48550/arXiv.1806.05394}\BibitemShut {NoStop}%
\bibitem [{\citenamefont {Erdmenger}\ \emph {et~al.}(2022)\citenamefont
  {Erdmenger}, \citenamefont {Grosvenor},\ and\ \citenamefont
  {Jefferson}}]{Erdmenger:2021sot}%
  \BibitemOpen
  \bibfield  {author} {\bibinfo {author} {\bibfnamefont {J.}~\bibnamefont
  {Erdmenger}}, \bibinfo {author} {\bibfnamefont {K.~T.}\ \bibnamefont
  {Grosvenor}}, \ and\ \bibinfo {author} {\bibfnamefont {R.}~\bibnamefont
  {Jefferson}},\ }\bibfield  {title} {\enquote {\bibinfo {title} {{Towards
  quantifying information flows: relative entropy in deep neural networks and
  the renormalization group}},}\ }\href {\doibase
  10.21468/SciPostPhys.12.1.041} {\bibfield  {journal} {\bibinfo  {journal}
  {SciPost Phys.}\ }\textbf {\bibinfo {volume} {12}},\ \bibinfo {pages} {041}
  (\bibinfo {year} {2022})},\ \Eprint {http://arxiv.org/abs/2107.06898}
  {arXiv:2107.06898 [hep-th]} \BibitemShut {NoStop}%
\bibitem [{\citenamefont {{Hernandez}}\ and\ \citenamefont
  {{Brown}}(2020)}]{2020arXiv200504305H}%
  \BibitemOpen
  \bibfield  {author} {\bibinfo {author} {\bibfnamefont {D.}~\bibnamefont
  {{Hernandez}}}\ and\ \bibinfo {author} {\bibfnamefont {T.~B.}\ \bibnamefont
  {{Brown}}},\ }\bibfield  {title} {\enquote {\bibinfo {title} {{Measuring the
  Algorithmic Efficiency of Neural Networks}},}\ }\href@noop {} {\bibfield
  {journal} {\bibinfo  {journal} {arXiv e-prints}\ ,\ \bibinfo {eid}
  {arXiv:2005.04305}} (\bibinfo {year} {2020})},\ \Eprint
  {http://arxiv.org/abs/2005.04305} {arXiv:2005.04305 [cs.LG]} \BibitemShut
  {NoStop}%
\bibitem [{\citenamefont {{Livni}}\ \emph {et~al.}(2014)\citenamefont
  {{Livni}}, \citenamefont {{Shalev-Shwartz}},\ and\ \citenamefont
  {{Shamir}}}]{2014arXiv1410.1141L}%
  \BibitemOpen
  \bibfield  {author} {\bibinfo {author} {\bibfnamefont {R.}~\bibnamefont
  {{Livni}}}, \bibinfo {author} {\bibfnamefont {S.}~\bibnamefont
  {{Shalev-Shwartz}}}, \ and\ \bibinfo {author} {\bibfnamefont
  {O.}~\bibnamefont {{Shamir}}},\ }\bibfield  {title} {\enquote {\bibinfo
  {title} {{On the Computational Efficiency of Training Neural Networks}},}\
  }\href@noop {} {\bibfield  {journal} {\bibinfo  {journal} {arXiv e-prints}\
  ,\ \bibinfo {eid} {arXiv:1410.1141}} (\bibinfo {year} {2014})},\ \Eprint
  {http://arxiv.org/abs/1410.1141} {arXiv:1410.1141 [cs.LG]} \BibitemShut
  {NoStop}%
\bibitem [{\citenamefont {{Hayou}}\ \emph {et~al.}(2018)\citenamefont
  {{Hayou}}, \citenamefont {{Doucet}},\ and\ \citenamefont
  {{Rousseau}}}]{2018arXiv180508266H}%
  \BibitemOpen
  \bibfield  {author} {\bibinfo {author} {\bibfnamefont {S.}~\bibnamefont
  {{Hayou}}}, \bibinfo {author} {\bibfnamefont {A.}~\bibnamefont {{Doucet}}}, \
  and\ \bibinfo {author} {\bibfnamefont {J.}~\bibnamefont {{Rousseau}}},\
  }\bibfield  {title} {\enquote {\bibinfo {title} {{On the Selection of
  Initialization and Activation Function for Deep Neural Networks}},}\
  }\href@noop {} {\bibfield  {journal} {\bibinfo  {journal} {arXiv e-prints}\
  ,\ \bibinfo {eid} {arXiv:1805.08266}} (\bibinfo {year} {2018})},\ \Eprint
  {http://arxiv.org/abs/1805.08266} {arXiv:1805.08266 [stat.ML]} \BibitemShut
  {NoStop}%
\bibitem [{\citenamefont {Hayou}\ \emph {et~al.}(2019)\citenamefont {Hayou},
  \citenamefont {Doucet},\ and\ \citenamefont {Rousseau}}]{pmlr-v97-hayou19a}%
  \BibitemOpen
  \bibfield  {author} {\bibinfo {author} {\bibfnamefont {S.}~\bibnamefont
  {Hayou}}, \bibinfo {author} {\bibfnamefont {A.}~\bibnamefont {Doucet}}, \
  and\ \bibinfo {author} {\bibfnamefont {J.}~\bibnamefont {Rousseau}},\
  }\bibfield  {title} {\enquote {\bibinfo {title} {On the impact of the
  activation function on deep neural networks training},}\ }in\ \href {\doibase
  https://doi.org/10.48550/arXiv.1902.06853} {\emph {\bibinfo {booktitle}
  {Proceedings of the 36th International Conference on Machine Learning}}},\
  \bibinfo {series} {Proceedings of Machine Learning Research}, Vol.~\bibinfo
  {volume} {97},\ \bibinfo {editor} {edited by\ \bibinfo {editor}
  {\bibfnamefont {K.}~\bibnamefont {Chaudhuri}}\ and\ \bibinfo {editor}
  {\bibfnamefont {R.}~\bibnamefont {Salakhutdinov}}}\ (\bibinfo  {publisher}
  {PMLR},\ \bibinfo {year} {2019})\ pp.\ \bibinfo {pages} {2672--2680},\
  \Eprint {http://arxiv.org/abs/1902.06853} {arXiv:1902.06853 [stat.ML]}
  \BibitemShut {NoStop}%
\bibitem [{\citenamefont {Roberts}\ \emph {et~al.}(2021)\citenamefont
  {Roberts}, \citenamefont {Yaida},\ and\ \citenamefont
  {Hanin}}]{Roberts:2021fes}%
  \BibitemOpen
  \bibfield  {author} {\bibinfo {author} {\bibfnamefont {D.~A.}\ \bibnamefont
  {Roberts}}, \bibinfo {author} {\bibfnamefont {S.}~\bibnamefont {Yaida}}, \
  and\ \bibinfo {author} {\bibfnamefont {B.}~\bibnamefont {Hanin}},\ }\bibfield
   {title} {\enquote {\bibinfo {title} {{The Principles of Deep Learning
  Theory}},}\ }\href {\doibase 10.1017/9781009023405} {\  (\bibinfo {year}
  {2021}),\ 10.1017/9781009023405},\ \Eprint {http://arxiv.org/abs/2106.10165}
  {arXiv:2106.10165 [cs.LG]} \BibitemShut {NoStop}%
\bibitem [{\citenamefont {Grosvenor}\ and\ \citenamefont
  {Jefferson}(2022)}]{Grosvenor:2021eol}%
  \BibitemOpen
  \bibfield  {author} {\bibinfo {author} {\bibfnamefont {K.~T.}\ \bibnamefont
  {Grosvenor}}\ and\ \bibinfo {author} {\bibfnamefont {R.}~\bibnamefont
  {Jefferson}},\ }\bibfield  {title} {\enquote {\bibinfo {title} {{The edge of
  chaos: quantum field theory and deep neural networks}},}\ }\href {\doibase
  10.21468/SciPostPhys.12.3.081} {\bibfield  {journal} {\bibinfo  {journal}
  {SciPost Phys.}\ }\textbf {\bibinfo {volume} {12}},\ \bibinfo {pages} {081}
  (\bibinfo {year} {2022})},\ \Eprint {http://arxiv.org/abs/2109.13247}
  {arXiv:2109.13247 [hep-th]} \BibitemShut {NoStop}%
\bibitem [{\citenamefont {Jefferson}(2020)}]{roCrit}%
  \BibitemOpen
  \bibfield  {author} {\bibinfo {author} {\bibfnamefont {R.}~\bibnamefont
  {Jefferson}},\ }\href@noop {} {\enquote {\bibinfo {title} {Criticality in
  deep neural nets},}\ }\bibinfo {howpublished}
  {\url{https://rojefferson.blog/2020/06/19/criticality-in-deep-neural-nets/}}
  (\bibinfo {year} {2020}),\ \bibinfo {note} {accessed: 2022-11-17}\BibitemShut
  {NoStop}%
\end{thebibliography}%

\end{document}